\documentclass[10pt,twocolumn,letterpaper]{article}

\usepackage[pagenumbers]{cvpr} 
\usepackage[table,dvipsnames]{xcolor}
\newcommand{\red}[1]{{\color{red}#1}}

\definecolor{cvprblue}{rgb}{0.21,0.49,0.74}
\usepackage[pagebackref,breaklinks,colorlinks,allcolors=cvprblue]{hyperref}

\def\paperID{2475} 
\def\confName{CVPR}
\def\confYear{2025}

\usepackage[utf8]{inputenc} 
\usepackage{CJKutf8} 
\usepackage{mathrsfs}
\usepackage{amsmath}
\usepackage{bm}
\usepackage{multirow}
\usepackage{bbding}
\usepackage{gensymb}

\usepackage{amsmath,amsfonts,bm}

\def\eqref#1{equation~\ref{#1}}

\def\1{\bm{1}}

\def\rvd{{\mathbf{d}}}

\def\rvf{{\mathbf{f}}}

\def\rvk{{\mathbf{k}}}

\def\rvo{{\mathbf{o}}}

\def\rvr{{\mathbf{r}}}

\def\rvv{{\mathbf{v}}}

\def\rvx{{\mathbf{x}}}

\DeclareMathAlphabet{\mathsfit}{\encodingdefault}{\sfdefault}{m}{sl}
\SetMathAlphabet{\mathsfit}{bold}{\encodingdefault}{\sfdefault}{bx}{n}

\DeclareMathOperator*{\argmin}{arg\,min}

\newcommand{\model}{\beps_\Theta}

\newcommand{\ldmencoderpose}{\mathcal{E}_\bpsi}
\newcommand{\ldmencoder}{\mathcal{E}_\bphi}

\newcommand{\bz}{{\mathbf{z}}}

\newcommand{\beps}{\boldsymbol{\epsilon}}

\newcommand{\btheta}{{\boldsymbol{\theta}}}
\newcommand{\bphi}{{\boldsymbol{\phi}}}
\newcommand{\bpsi}{{\boldsymbol{\psi}}}

\newcommand{\bzero}{\mathbf{0}}
\newcommand{\beye}{\mathbf{I}}

\newcommand{\N}{\mathcal{N}}

\newcommand{\set}[1]{\mathcal{#1}}

\newcommand{\loss}{\mathcal{L}}

\usepackage[dvipsnames]{xcolor}
\definecolor{magenta(process)}{rgb}{1.0, 0.0, 0.9}
\newcommand{\heading}[1]{\noindent\textbf{#1.}}

\newcommand{\nickname}{\NICKNAME} 

\newcommand{\real}{\mathbb{R}}

\newcommand{\RN}[1]{%
  \textup{\uppercase\expandafter{\romannumeral#1}}%
}

\definecolor{yellow}{rgb}{1, 1, 0.7}
\definecolor{orange}{rgb}{1, 0.85, 0.7}
\definecolor{tablered}{rgb}{1, 0.7, 0.7}
\definecolor{red}{rgb}{1, 0, 0}

\newcommand{\blue}[1]{{\textcolor{blue}{#1}}}
\definecolor{rred}{RGB}{245, 152, 153}
\definecolor{oorange}{RGB}{253, 205, 154}

\newcommand{\NICKNAME}{\textsc{3DEnhancer}}

\title{\nickname{}: Consistent Multi-View Diffusion for 3D Enhancement}

\author{Yihang Luo \quad Shangchen Zhou$^{\dag}$ \quad Yushi Lan \quad Xingang Pan \quad Chen Change Loy$^{\dag}$\\
S-Lab, Nanyang Technological University \\
{\tt\small \url{https://yihangluo.com/projects/3DEnhancer}}
}

\begin{document}
\twocolumn[{%
\renewcommand\twocolumn[1][]{#1}%
\maketitle
\vspace{-2mm}
\begin{center}
    \centering
    \includegraphics[width=\linewidth]{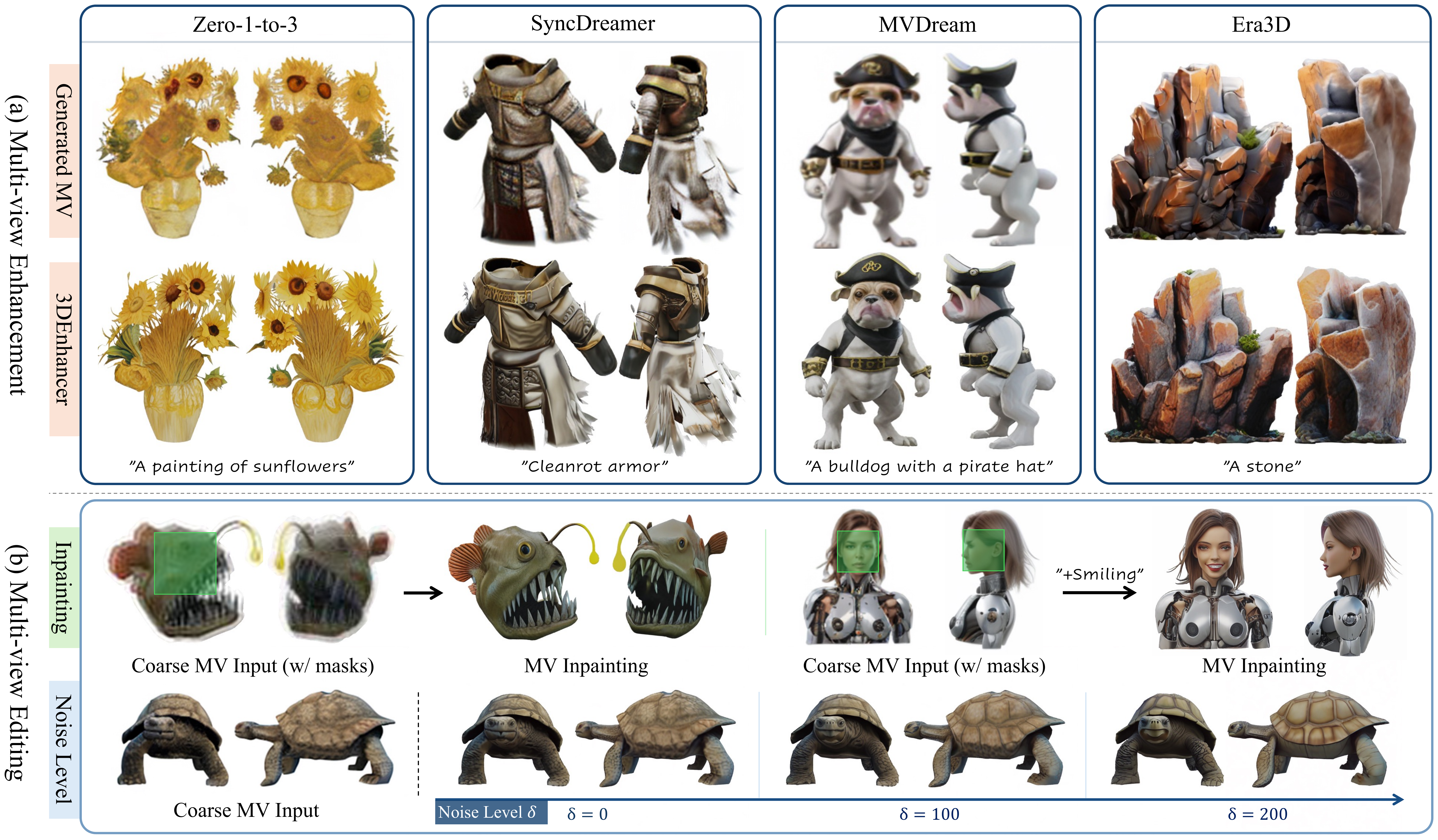}
    \vspace{-5mm}
    \captionof{figure}{
    Our proposed 3DEnhancer showcases excellent capabilities in enhancing multi-view images generated by various models. As shown in (a), it can significantly improve texture details, correct texture errors, and enhance consistency across views. Beyond enhancement, as illustrated in (b), 3DEnhancer also supports texture-level editing, including regional inpainting, and adjusting texture enhancement strength via noise level control.
    \textbf{(Zoom-in for best view)}
    } \vspace{2mm}
    \label{fig:teaser}
\end{center}%
}]

\def\thefootnote{\dag}\footnotetext{Corresponding authors.}

\begin{abstract}
Despite advances in neural rendering, due to the scarcity of high-quality 3D datasets and the inherent limitations of multi-view diffusion models, view synthesis and 3D model generation are restricted to low resolutions with suboptimal multi-view consistency.
In this study, we present a novel 3D enhancement pipeline, dubbed \nickname{}, which employs a multi-view latent diffusion model to enhance coarse 3D inputs while preserving multi-view consistency. 
Our method includes a pose-aware encoder and a diffusion-based denoiser to refine low-quality multi-view images, along with data augmentation and a multi-view attention module with epipolar aggregation to maintain consistent, high-quality 3D outputs across views. Unlike existing video-based approaches, our model supports seamless multi-view enhancement with improved coherence across diverse viewing angles. Extensive evaluations show that \nickname{} significantly outperforms existing methods, boosting both multi-view enhancement and per-instance 3D optimization tasks.
\end{abstract}
\vspace{-8mm}    
\section{Introduction}
\label{sec:intro}
The advancements in generative models~\cite{Jo2022DDPM,Goodfellow2014GenerativeAN} and differentiable rendering~\cite{mildenhall2020nerf} have paved the way for a new research field known as neural rendering~\cite{Tewari2021AdvancesIN}. In addition to pushing the boundaries of view synthesis~\citep{kerbl3Dgaussians}, the generation and editing of 3D models~\cite{3dshape2vecset,shi2023MVDream,Jun2023ShapEGC,lan2024ln3diff,zhang2024clay,liu2023zero1to3,li2024era3d,lan2024ga,chen2024sar3d,hoellein2024viewdiff} has become achievable. These methods are trained on the large-scale 3D datasets, \eg, Objaverse dataset~\cite{objaverse}, enabling fast and diverse 3D synthesis.

Despite these advances, several challenges remain in 3D generation. A key limitation is the scarcity of high-quality 3D datasets; unlike the billions of high-resolution image and video datasets available~\citep{Schuhmann2022LAION5BAO}, current 3D datasets~\cite{objaverseXL} are limited to a much smaller scale~\cite{qiu2024richdreamer}.
Another limitation is the dependency on multi-view (MV) diffusion models~\cite{shi2023MVDream,shi2023zero123plus}. Most state-of-the-art 3D generative models~\cite{tang2024lgm,wang2024crm} follow a \textit{two-stage} pipeline: first, generating multi-view images conditioned on images or text~\citep{wang2023imagedream,shi2023MVDream}, and then reconstructing 3D models from these generated views~\cite{hong2023lrm,tang2024lgm}. Consequently, the low-quality results and view inconsistency issues of multi-view diffusion models~\cite{shi2023MVDream} restrict the quality of the final 3D output.
Besides, existing novel view synthesis methods~\cite{mildenhall2020nerf, kerbl3Dgaussians} usually require dense, high-resolution input views for optimization, making 3D content creation challenging when only low-resolution sparse captures are available.

In this study, we address these challenges by introducing a versatile 3D enhancement framework, dubbed~\nickname{}, which leverages a text-to-image diffusion model as the 2D generative prior to enhance general coarse 3D inputs. The core of our proposed method is a multi-view latent diffusion model (LDM)~\cite{rombach2022LDM} designed to enhance coarse 3D inputs while ensuring multi-view consistency.
Specifically, the framework consists of a pose-aware image encoder that encodes low-quality multi-view renderings into latent space and a multi-view-based diffusion denoiser that refines the latent features with view-consistent blocks.
The enhanced views are then either used as input for multi-view reconstruction or directly serve as reconstruction targets for optimizing the coarse 3D inputs. 

To achieve practical results, we introduce diverse degradation augmentations~\cite{wang2021realesrgan} to the input multi-view images, simulating the distribution of coarse 3D data. In addition, we incorporate efficient multi-view row attention~\cite{li2024era3d,huang2023epidiff} to ensure consistency across multi-view features. To further reinforce coherent 3D textures and structures under significant viewpoint changes, we also introduce near-view epipolor aggregation modules, which directly propagate corresponding tokens across near views using epipolar-constrained feature matching~\cite{tokenflow2023,chen2024dge}. These carefully designed strategies effectively contribute to achieving high-quality, consistent multi-view enhancement.

The most relevant works to our study are 3D enhancement approaches using video diffusion models~\cite{Shen2024SuperGaussian,liu20243dgsenhancer}. While video super-resolution (SR) models~\cite{xu2024videogigagan} can also be adapted for 3D enhancement, several challenges that make them less suitable for use as generic 3D enhancers. First, these methods are limited to enhancing 3D model reconstructions through per-instance optimization, whereas our approach can seamlessly enhance 3D outputs by integrating multi-view enhancement into the existing \textit{two-stage} 3D generation frameworks (\eg, from ``MVDream~\cite{shi2023MVDream} $\to$ LGM~\cite{tang2024lgm}'' to ``MVDream $\to$ \nickname{} $\to$ LGM''). Second, video models often struggle with long-term consistency and fail to correct generation artifacts in 3D objects under significant viewpoint variations. Besides, video diffusion models based on temporal attention~\cite{svd} face limitations in handling long videos due to memory and speed constraints. In contrast, our multi-view enhancer models texture correspondences across various views both implicitly and explicitly, by utilizing multi-view row attention and near-view epipolar aggregation, leading to superior view consistency and higher efficiency.

In summary, we present a novel \nickname{} for generic 3D enhancement using multi-view denoising diffusion. Our contributions include a robust data augmentation pipeline, and the hybrid view-consistent blocks that integrate multi-view row attention and near-view epipolar aggregation modules to promote view consistency. Compared to existing enhancement methods, our multi-view 3D enhancement framework is more versatile and supports texture refinement.
We conduct extensive experiments on both multi-view enhancement and per-instance optimization tasks to evaluate the model's components. Our proposed pipeline significantly improves the quality of coarse 3D objects and consistently surpasses existing alternatives.
\section{Related Work}
\label{sec:relatedwork}
\heading{3D Generation with Multi-view Diffusion}
The success of 2D diffusion models ~\citep{song2021scorebased,Jo2022DDPM} has inspired their application to 3D generation. Score distillation sampling (SDS)~\citep{poole2022dreamfusion,wang2023prolificdreamer} distills 3D from a 2D diffusion model but faces challenges like expensive optimization, mode collapse, and the Janus problem.
More recent methods propose learning the 3D via a two-stage pipeline: multi-view images generation~\citep{shi2023MVDream,long2023wonder3d,shi2023zero123plus,wang2024crm} and feed-forward 3D reconstruction~\citep{hong2023lrm,xu2024instantmesh,tang2024lgm}.
Though yielding promising results, their performance is bounded by the quality of the multi-view generative models, including the violation of strict view consistency~\citep{liu2023zero1to3} and failing to scale up to higher resolution~\citep{shi2023zero123plus}.
Recent work has focused on developing more 3D-aware attention operations, such as epipolar attention~\cite{tseng2023consistent,huang2023epidiff} and row-wise attention~\cite{li2024era3d}. However, we find that enforcing strict view consistency remains challenging when relying solely on attention-based operations.

\heading{Image and Video Super-Resolution}
Image and video SR aim to improve visual quality by upscaling low-resolution content to high resolution. Research in this field has evolved from focusing on pre-defined single degradations~\cite{Zhang_2018_ECCV, wang2018esrgan, zhou2020cross, liang2021swinir, chen2021pre, chen2023activating, wang2019edvr, chan2021basicvsr, chan2022basicvsr++, liang2022rvrt} (\eg, bicubic downsampling) to addressing unknown and complex degradations~\cite{zhang2021designing, wang2021realesrgan, chan2022realbasicvsr} in real-world scenarios. To tackle real-world enhancement, some studies~\cite{zhang2021designing, wang2021realesrgan, chan2022realbasicvsr, zhou2022lednet} introduce effective degradation pipelines that simulate diverse degradations for data augmentation during training, significantly boosting performance in handling real-world cases. To achieve photorealistic enhancement, recent work has integrated various generative priors to produce detailed textures, including StyleGAN~\cite{chan2021glean, wang2021gfpgan, Yang2021GPEN}, codebook~\cite{zhou2022codeformer, chen2023iterative}, and the latest diffusion models~\cite{wang2024stablesr, zhou2024upscale}. For instance, StableSR~\cite{wang2024stablesr} leverages the pretrained image diffusion model, \ie, Stable Diffusion (SD)~\cite{rombach2022LDM}, for image enhancement, while Upscale-A-Video~\cite{zhou2024upscale} further extends the diffusion model for video upscaling. Video SR networks commonly employ recurrent frame fusion~\cite{zhou2019stfan, wang2019edvr}, optical flow-guided propagation~\cite{chan2021basicvsr, chan2022basicvsr++, chan2022realbasicvsr, liang2022rvrt} or temporal attention~\cite{zhou2024upscale} to enhance temporal consistency across adjacent frames. However, due to large spatial misalignments from viewpoint changes, these methods face challenges in establishing long-range correspondences across multi-view images, making them unsuitable for multi-view fusion for 3D. In this study, we focus on exploiting a image diffusion model to achieve robust 3D enhancement while preserving view consistency.

\heading{3D Texture Enhancement}
With the rapid advancement of 3D generative models~\cite{zhang2024clay,bensadoun2024meta,lan2024ln3diff,lan2024ga,chen2024sar3d}, attention is paid to further improve 3D generation quality through a cascade 3D enhancement module. 
%
Meta 3D Gen~\cite{bensadoun2024meta,Bensadoun2024Meta3T} proposes a UV space enhancement model to achieve sharper textures. However, training the UV-specific enhancement model requires spatially continuous UV maps, which are limited in both quantities~\cite{objaverse} and qualities~\cite{xatlas}.
Intex~\cite{tang2024intex} and SyncMVD~\cite{liu2023text} also employ UV space for generating and enhancing 3D textures. However, these techniques are specifically designed for 3D mesh with UV coordinates, making them unsuitable for other 3D representations like 3DGS~\cite{kerbl3Dgaussians}.
Unique3D~\cite{wu2024unique3d} and CLAY~\cite{zhang2024clay} apply 2D enhancement module RealESRGAN~\cite{wang2021realesrgan} directly to the generated multi-view outputs. Though straightforward, this approach risks compromising 3D consistency across the multi-view results.
MagicBoost~\cite{yang2024magicboost} introduces a 3D refinement pipeline but relies on computationally expensive SDS optimization. Deceptive-NeRF/3DGS~\cite{liu2024deceptive} uses an image diffusion model to generate high-quality pseudo-observations for novel views but requires a few accurately captured sparse views as key inputs. SuperGaussian~\cite{Shen2024SuperGaussian} and 3DGS-Enhancer~\cite{liu20243dgsenhancer} propose to enhance 3D through 2D video generative priors~\cite{svd,xu2024videogigagan}. These pre-trained video models struggle to maintain long-range consistency under large viewpoint variations, making them less effective at fixing texture errors in multi-view generation.

\section{Methodology}
\label{sec:method}
\begin{figure*}[t]
\begin{center}
    \includegraphics[width=.99\linewidth]{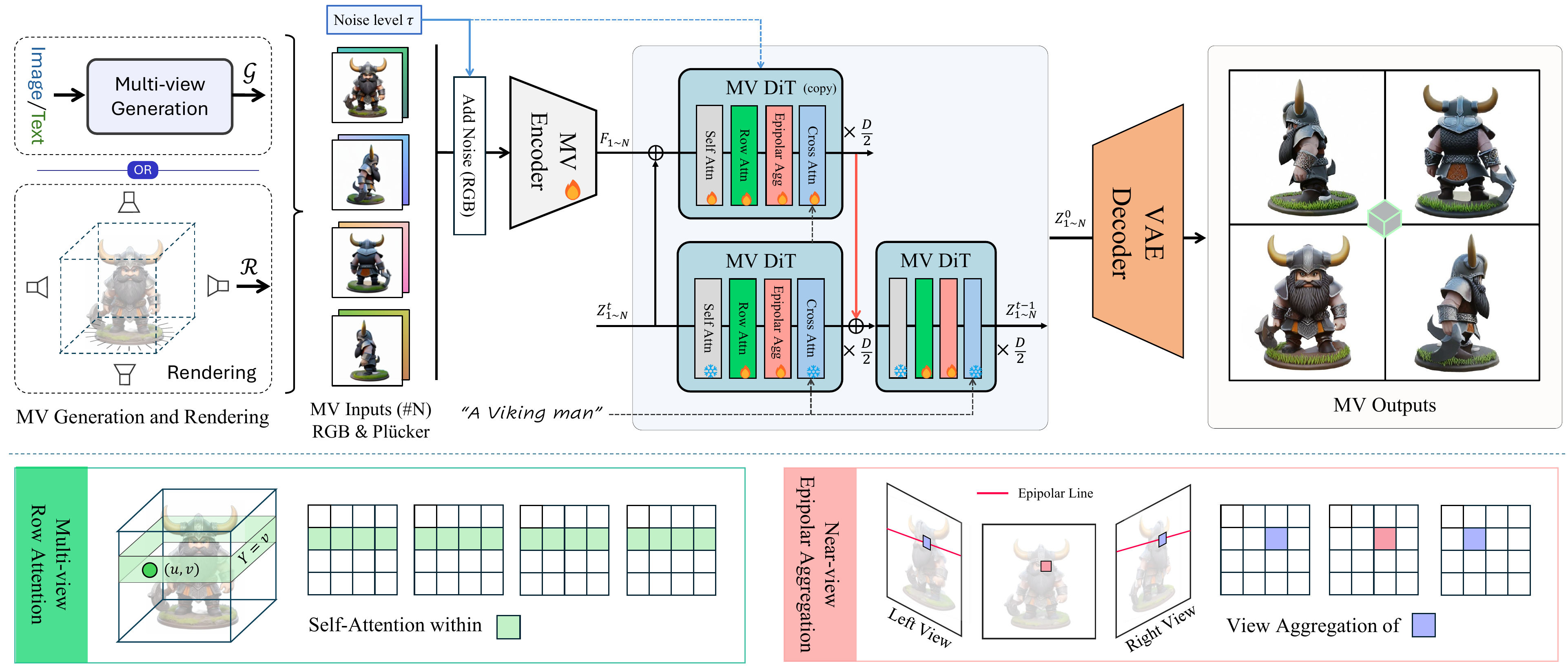}
    \vspace{-1mm}
    \caption{
    An overview of~\nickname{}. By harnessing generative priors, \nickname{} adapts a text-to-image diffusion model to a multi-view framework aimed at 3D enhancement. It is compatible with multi-view images generated by models like \text{MVDream}~\cite{shi2023MVDream} or those rendered from coarse 3D representations like NeRFs~\cite{mildenhall2020nerf} and 3DGS~\cite{kerbl3Dgaussians}. Given LQ multi-view images along with their corresponding camera poses, \nickname{} aggregates multi-view information within a DiT~\cite{Peebles2022DiT} framework using row attention and epipolar aggregation modules, improving visual quality while preserving consistency across views. Furthermore, the model supports texture-level editing via text prompts and adjustable noise levels, allowing users to correct texture errors and control the enhancement strength.
    }
    \vspace{-6mm}
\label{fig:overview}
\end{center}
\end{figure*}

A common pipeline in current 3D generation involves an \textit{image-to-multiview} stage~\cite{wang2023imagedream}, followed by \textit{multiview-to-3D}~\cite{tang2024lgm} generation that converts these multi-view images into a 3D object. However, due to limitations in resolution and view consistency~\cite{liu2023zero1to3}, the resulting 3D outputs often lack high-quality textures and detailed geometry. The proposed multi-view enhancement network, \nickname{}, aims at improving the quality of 3D representations. Our motivation is that if we can obtain high-quality and view-consistent multi-view images, then the quality of 3D generation can be correspondingly enhanced.

As illustrated in Fig.~\ref{fig:overview}, our framework employs a Diffusion Transformer (DiT) based LDM~\cite{Peebles2022DiT,chen2024pixartsigma} as the backbone. We incorporate a pose-aware encoder and view-consistent DiT blocks to ensure multi-view consistency, allowing us to leverage the powerful multi-view diffusion models to enhance both coarse multi-view images and 3D models. The enhanced multi-view images can improve the performance of pre-trained feed-forward 3D reconstruction models, \eg, LGM~\cite{tang2024lgm}, as well as optimize a coarse 3D model through iterative updates.

\heading{Preliminary: Multi-view Diffusion Models}
LDM~\cite{rombach2022LDM,vahdat2021score,Jo2022DDPM} is designed to acquire a prior distribution $p_\btheta(\bz)$ within the perceptual latent space, whose training data is the latent obtained from the trained VAE encoder $\ldmencoder$.
By training to predict a denoised variant of the noisy input $\bz_t=\alpha_t \bz + \sigma_t \bm{\epsilon}$ at each diffusion step $t$, $\model$ gradually learns to denoise from a standard Normal prior $\N(\bzero, \beye)$ by solving a reverse SDE~\cite{Jo2022DDPM}.

Similarly, multi-view diffusion generation models~\cite{xu2023dmv3d,shi2023MVDream} consider the joint distribution of multi-view images $\set{X}=\{\rvx_1,\ldots,\rvx_N\}$, where each set of $\set{X}$ contains RGB renderings $\rvx_{\rvv} \in \real^{H \times W \times 3}$ from the same 3D asset given viewpoints $\set{C}=\{\pi_1, \ldots, \pi_N\}$. The latent diffusion process is identical to diffusing each encoded latent $\bz =\ldmencoder(\rvx)$ independently with the shared noise schedule: $\set{Z}_t=\{\alpha_t \bz + \sigma_t \bm{\epsilon} \mid \bz \in \set{Z}\}$. Formally, given the multi-view data $\mathcal{D}_{mv} := \{\set{X}, \set{C}, y\}$, the corresponding diffusion loss is defined as:
\begin{equation}
\loss_{MV}(\theta, \mathcal{D}_{mv})=
\mathbb{E}_{\set{Z}, y, \pi,t,\bm{\epsilon}}\left[ \| \epsilon -\model(\set{Z}_t; y,\pi,t)\|_2^2\right], 
\label{eq:denoising_objective}
\end{equation}
where $y$ is the optional text or image condition.

\subsection{Pose-aware Encoder}
Given the posed multi-view images~$\set{X}$, we add controllable noise to the images as an augmentation to enable controllable refinement, as described later in Sec.~\ref{sec:3.3}.
To further inject camera condition for each view $v$, we follow the prior work~\cite{sitzmann2021light, xu2023dmv3d, lan2024ln3diff, tang2024lgm}, and concatenate Plücker coordinates $\rvr_\rvv^i = (\rvd^i, \rvo^i \times \rvd^i) \in \real^{6}$ with image RGB values $\rvx_{\rvv}^i \in \real^{3}$ along the channel dimension. Here, $\rvo^i$ and $\rvd^i$ are the ray origin and ray direction for pixel $i$ from view $\rvv$, and $\times$ denotes the cross product. We then send the concatenated results to a trainable pose-aware multi-view encoder $\ldmencoderpose$, whose outputs are injected into the pre-trained DiT through a learnable copy~\cite{zhang2023adding}. 


\subsection{View-Consistent DiT Block}
The main challenge of 3D enhancement is achieving precise view consistency across generated 2D multi-view images. 
Multi-view diffusion methods commonly rely on multi-view attention layers to exchange information across different views, aiming to generate multiview-consistent images. A prevalent approach is extending self-attention to all views, known as dense multi-view attention~\cite{shi2023MVDream,long2023wonder3d}. While effective, this method significantly raises both computational demands and memory requirements. To further enhance the effectiveness and efficiency of inter-view aggregation, we introduce two efficient modules in the DiT blocks: multi-view row attention and near-view epipolar aggregation, as shown in Fig.~\ref{fig:overview}.

\heading{Multi-view Row Attention}
To enhance the noisy input views to higher resolution, \eg, $512 \times 512$, efficient multi-view attention is required to facilitate cross-view information fusion. Considering the epipolar constraints~\cite{Hartley2004}, the 3D correspondences across views always lie on the epipolar line~\cite{tseng2023consistent,huang2023epidiff}.
Since our diffusion denoising is performed on $16\times$ downsampled features~\cite{chen2024pixartsigma}, and typical multi-view settings often involve elevation angles around $0^\circ$, we assume that horizontal rows approximate the epipolar line. Therefore, we adopt the special epipolar attention, specifically the multi-view row attention~\cite{li2024era3d}, enabling efficient information exchange among multi-view features.

Specifically, the input cameras are chosen to look at the object with their $Y$ axis aligned with the gravity direction and cameras' viewing directions are approximately horizontal (i.e., the pitch angle is generally level, with no significant deviation). This case is visualized in Fig.~\ref{fig:overview}, for a coordinate $(u, v)$ in the attention feature space of one view, the corresponding epipolar line in the attention feature space of other views can be approximated as $Y = v$. This enables the extension of self-attention layers calculated on tokens within the same row across multiple views to learn 3D correspondences. As ablated in Tab.~\ref{tab:ablation}, the multi-view row attention can efficiently encourage view consistency with minor memory consumption.

\heading{Near-view Epipolar Aggregation}
Though multi-view attention can effectively facilitate view consistency, we observe that the attention-only operation still struggles with accurate correspondences across views. To address this issue, we incorporate explicit feature aggregation among neighboring views to ensure multi-view consistency. Specifically, given the output features $\{\rvf_\rvv\}_{\rvv=1}^{N}$ from the multi-view row attention layers for each posed multi-view input, we propagate features by finding near-view correspondences with epipolar line constraints.
Formally, for the feature map $\rvf_{\rvv}$ corresponding to the posed image $\rvx_{\rvv}$, we calculate its correspondence map $M_{\rvv, \rvk}$ with the near views $\rvk$ as follows:
\begin{equation}
    M_{\rvv, \rvk}[i] = \argmin_{j, \ j{^\top} Fi = 0} D(\rvf_\rvv[i], \rvf_{\rvk}[j]), 
\end{equation}
where $D$ denotes the cosine distance, and $\rvk \in \{\rvv-1, \rvv+1\}$ represents the two nearest neighbor views of the given pose. Here, $i$ and $j$ are indices of the spatial locations in the feature maps, $F$ is the fundamental matrix relating the two views $\rvv$ and $\rvk$, and the index $j$ lies on the epipolar line in the view $\rvk$, subject to the constraint $j^{\top} F i = 0$. We then obtain the aggregated feature map $\widetilde{\rvf}_\rvv$ of the view $\rvv$ by linearly combining features of correspondences from the two nearest views via:
\begin{equation}
\begin{aligned}
\widetilde{\rvf}_\rvv[i] &= w \cdot \rvf_{\rvv{-}1}[M_{\rvv, \rvv-1}[i]] \\
&\quad + (1 - w) \cdot \rvf_{\rvv{+}1}[M_{\rvv, \rvv+1}[i]],
\end{aligned}
\label{eq:aggregate_view}
\end{equation}
where $w$ represents the weight to combine the features of the two nearest views. The calculation of $w$ uses a hybrid fusion strategy, which ensures that the weight assignment accounts for both the \textit{physical camera distance} and the \textit{token feature similarity} (see the Appendix Sec.~\ref{sec:weight}). As the feature aggregation process is non-differentiable, we adopt the straight-through estimator $\text{sg}[\cdot]$ in VQVAE~\cite{Oord2017NeuralDR} to facilitate gradient back-propagation in the token space. 
Near-view epipolar aggregation explicitly propagates tokens from neighboring views, which greatly improves view consistency. However, due to substantial view changes, the corresponding tokens may not be available, leading to unexpected artifacts during token replacement. To address this, we fuse the feature $\rvf_\rvv$ of the current view with the feature $\widetilde{\rvf}_\rvv$ from near-view epipolar aggregation, with 0.5 averaging. This effectively combines multi-view row attention and near-view epipolar aggregation, thereby enhancing view consistency both implicitly and explicitly.

This approach is similar to token-space editing methods like TokenFlow~\cite{tokenflow2023} and DGE~\cite{chen2024dge}. However, we propose a trainable version that considers both geometric and feature similarity for effective feature fusion.

\subsection{Multi-view Data Augmentation}
\label{sec:3.3}
Our goal is to train a versatile and robust enhancement model that performs well on low-quality multi-view images from diverse data sources, such as those generated by image-to-3D models or rendered from coarse 3D representations. To achieve this, we carefully design a comprehensive data augmentation pipeline to expand the distribution of distortions in our base training data, bridging the domain gap between training and inference.

\begin{figure*}[ht]
\vspace{-3mm}
\begin{center}
    \includegraphics[width=.99\linewidth]{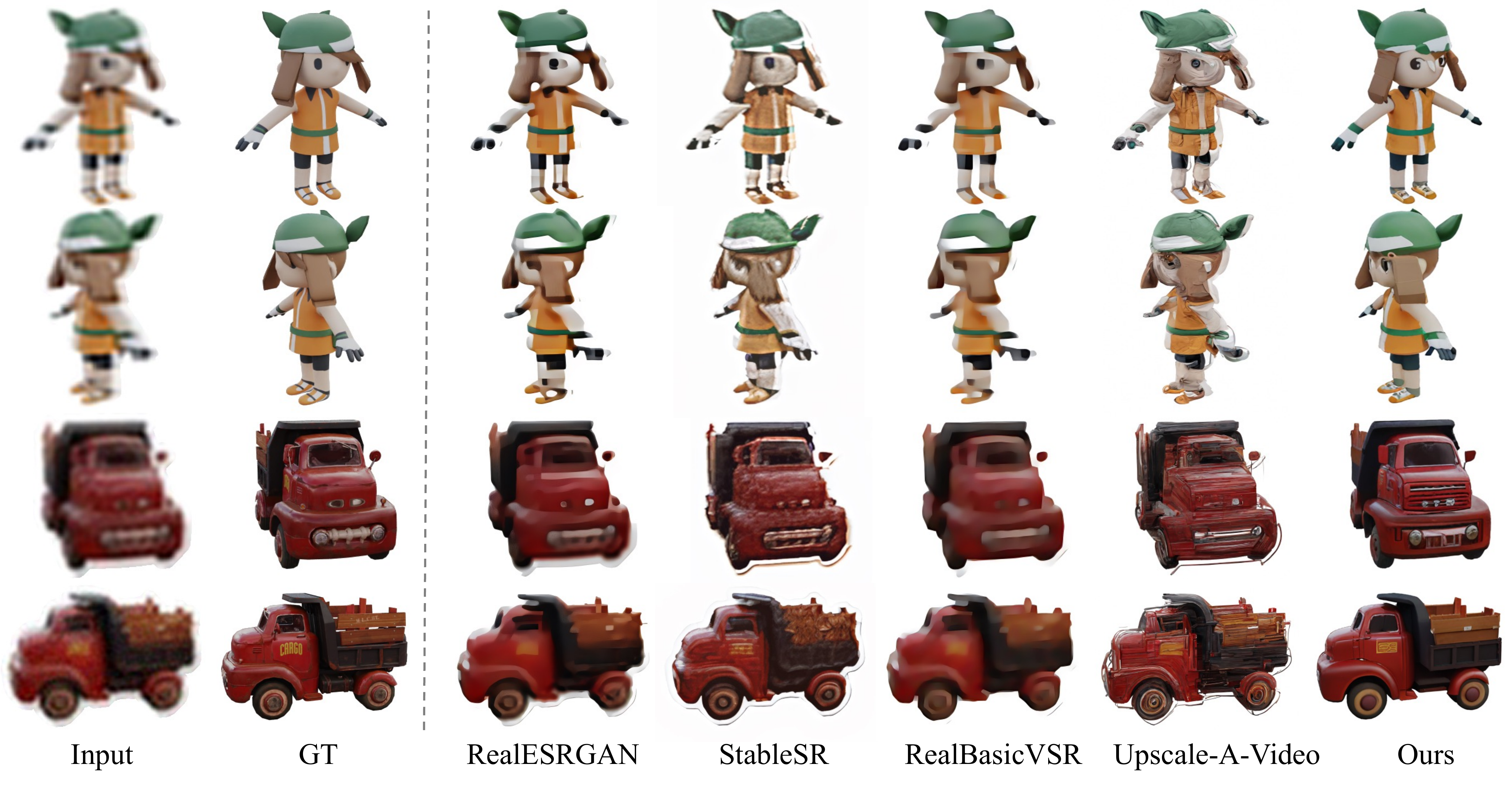}
    \vspace{-3mm}
    \caption{
    Qualitative comparisons of enhancing  multi-view synthesis on the Objaverse synthetic dataset. As can be seen, only \nickname{} can correct flowed and missing textures with view consistency.
    }
    \vspace{-6mm}
\label{fig:qualitative_synthetic}
\end{center}
\end{figure*}

\heading{Texture Distortion}
To emulate the low-quality textures and local inconsistencies found in synthesized multi-view images, we employ a texture degradation pipeline commonly used in 2D enhancement~\cite{wang2021realesrgan, zhou2024upscale}. This pipeline randomly applies downsampling, blurring, noise, and JPEG compression to degrade the image quality. 

\heading{Texture Deformation and Camera Jitter}
As in LGM~\cite{tang2024lgm}, we introduce grid distortion to simulate texture inconsistencies in multi-view images and apply camera jitter augmentation to introduce variations in the conditional camera poses of multi-view inputs.

\heading{Color Shift}
We also observe color variations in corresponding regions between multi-view images generated by image-to-3D models. By randomly applying color changes to some image patches, we encourage the model to produce results with consistent colors. 
%
In addition, renderings from a coarse 3DGS sometimes result in a grayish overlay or ghosting artifacts, akin to a translucent mask. To simulate this effect, we randomly apply a semi-transparent object mask to the image, allowing the model to learn to remove the overlay and improve 3D visual quality.

\heading{Noise-level Control}
To control the enhancement strength, we apply noise augmentation by adding controllable noise to the input multi-view images. This noise augmentation process is similar to the diffusion process in diffusion models. This approach can further enhance the model's robustness in handling unseen artifacts~\cite{zhou2024upscale}. 


\subsection{Inference for 3D Enhancement}

We present two ways to utilize our \nickname{} for 3D enhancement:
\begin{itemize}
    \item The proposed method can be directly applied to generation results from existing multi-view diffusion models~\cite{shi2023MVDream, liu2023zero1to3, liu2023syncdreamer, eva3d}, and the enhanced output shall serve as the input to the multi-view 3D reconstruction models~\cite{tang2024lgm,openlrm,wang2024crm,xu2024grm}. Given the enhanced multi-view inputs with sharper textures and view-consistent geometry, our method can be directly used to improve the quality of existing multi-view to 3D reconstruction frameworks.
    \item Our method can also be used for \emph{directly} enhancing a coarse 3D model through iterative optimization.
    Specifically, given an initial coarse 3D reconstruction as $\mathcal{M}$ and a set of viewpoints $\{\pi_\rvv\}_{\rvv=1}^{N}$, we first render the corresponding views $\set{X} = \{\rvx_\rvv\}_{\rvv=1}^N$ where $\rvx_\rvv = \text{Rend}(\mathcal{M}, \pi_\rvv)$ is obtained with the corresponding rendering techniques~\cite{mildenhall2020nerf,kerbl3Dgaussians}. Let $\set{X}^{\prime}=\{\rvx_\rvv^\prime\}_{\rvv=1}^N$ be the enhanced multi-view images, we can then update the 3D model $\mathcal{M}$ by supervising it with $\set{X}^{\prime}$ as 
    %
    %
    \begin{equation}
       \mathcal{M}^{\prime} = \argmin_{\mathcal{M}}\sum_{\rvv=1}^{N} \loss({\rvx_\rvv^\prime, \text{Rend}(\mathcal{M}, \pi_\rvv))}.
    \end{equation}
    Following previous methods that reconstruct 3D from synthesized 2D images~\cite{wu2023reconfusion,gao2024cat3d}, we use a mixture of $\loss_\text{1}$ and $\loss_\text{LPIPS}$~\cite{zhang2018unreasonable} for robust optimization. In practice, unlike iterative dataset updates (IDU)~\cite{instructnerf2023}, we found that inferring the enhanced views $\set{X}^{\prime}$ once already yields high-quality results. More implementation details and results for this part are provided in the Appendix Sec.~\ref{sec:optimization}.
\end{itemize}

\section{Experiments}
\subsection{Datasets and Implementation}

\noindent {\bf Datasets.}
For training, we use the Objaverse dataset \cite{objaverse}, specifically leveraging the G-buffer Objaverse \cite{qiu2024richdreamer}, which provides diverse renderings on Objaverse instances. We construct LQ-HQ view pairs following the augmentation pipeline outlined in \hyperref[sec:3.3]{Sec.~3.3} and then split the dataset into separate training and test sets. Overall, approximately $400$K objects are used for training. For each object, we randomly sample 4 input views with azimuth angles ranging from $0^\circ$ to $360^\circ$ and elevation angles between $-5^\circ$ and $30^\circ$.

For evaluation, we use a test set containing $500$ objects from different categories within our synthesized Objaverse datasets. We further evaluate our model on the zero-shot in-the-wild dataset by selecting images from the GSO dataset \cite{gso}, image diffusion model outputs \cite{rombach2022LDM}, and web-sourced content. These images are then processed using several novel view synthesis methods \cite{liu2023zero1to3, shi2023MVDream, liu2023syncdreamer, li2024era3d} to create our in-the-wild test set, containing a total of 400 instances.

\noindent {\bf Implementation Details.} 
We employ PixArt-$\Sigma$~\cite{chen2024pixartsigma}, an efficient DiT model, as our backbone. Our model is trained on images with a resolution of 512 x 512. The AdamW optimizer \cite{Kingma2015AdamAM} is used with a fixed learning rate of 2e-5. Our training is conducted over 10 days using 8 Nvidia A100-80G GPUs, with a batch size 256. For inference, we employ a DDIM sampler \cite{song2020denoising} with 20 steps and set the Classifier-Free Guidance (CFG)~\cite{Ho2022ClassifierFreeDG} scale to 4.5.

\begin{figure*}[ht]
\begin{center}
\includegraphics[width=\linewidth]{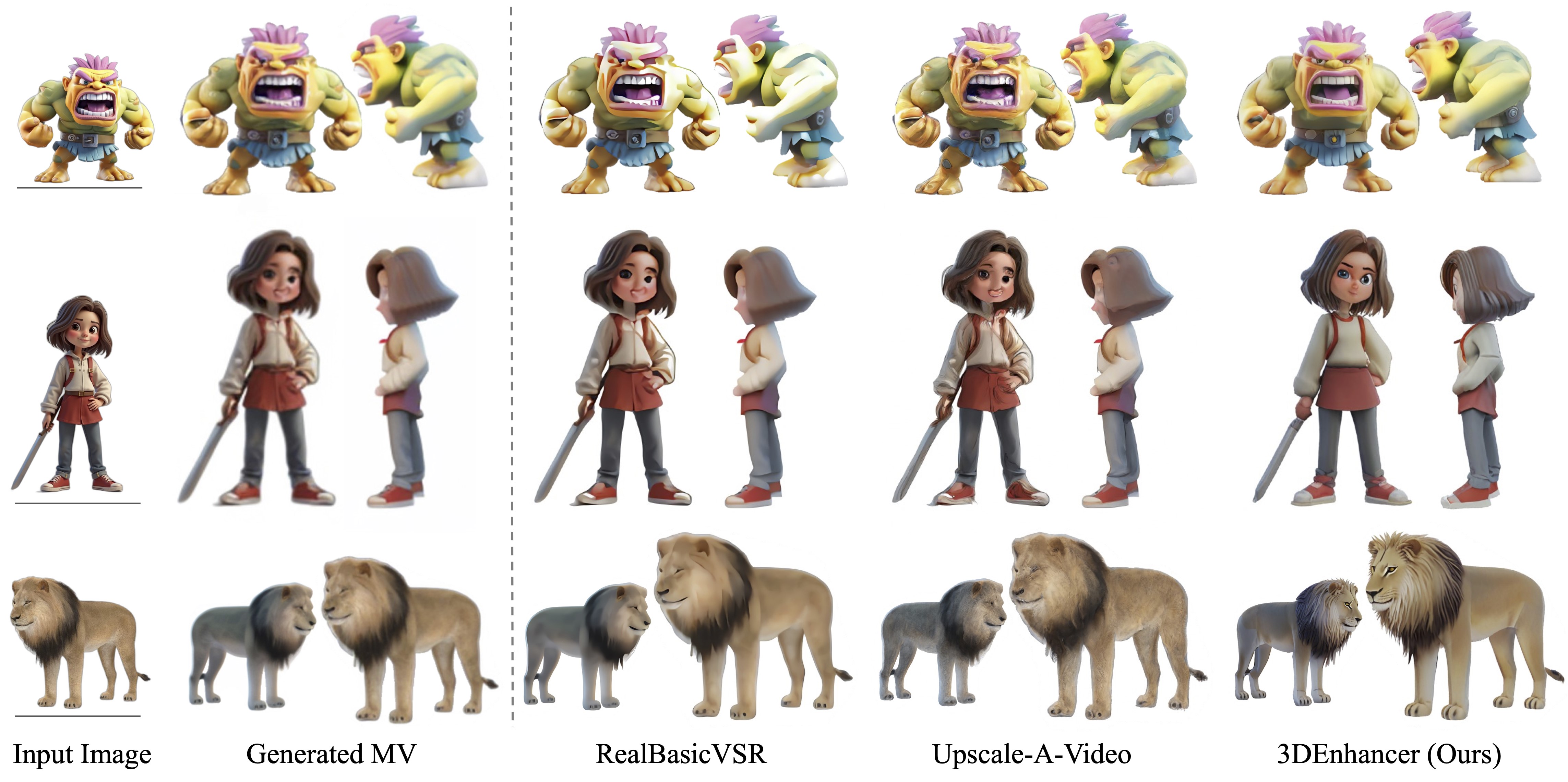}
    \vspace{-5mm}
    \caption{
    Qualitative comparisons of enhancing  multi-view synthesis with RealBasicVSR\cite{chan2022realbasicvsr} and Upscale-A-Video\cite{zhou2024upscale} on the in-the-wild dataset. Visually inspecting, \nickname{} yields sharp and consistent textures with intact semantics, such as the eyes of the girl.
    }
    \vspace{-5mm}
\label{fig:qualitive_inthewild}
\end{center}
\end{figure*}

\noindent {\bf Baselines.}
To assess the effectiveness of our approach, we adopt two image enhancement models, RealESRGAN \cite{wang2021realesrgan} and StableSR \cite{wang2024stablesr}, along with two video enhancement models, RealBasicVSR \cite{chan2022realbasicvsr} and Upscale-a-Video \cite{zhou2024upscale} as our baselines. For a fair comparison, we further fine-tune all these methods on the Objaverse dataset to minimize potential data domain discrepancies. During inference, since Real-ESRGAN, RealBasicVSR, and Upscale-a-Video by default produce images upscaled by a factor of ×4, we resize their outputs to a uniform resolution of 512 × 512 for comparison.

\noindent {\bf Metrics.}
We evaluate the effectiveness of our methods on two tasks: multi-view synthesis enhancement and 3D reconstruction improvement. We employ standard metrics including PSNR, SSIM, and LPIPS \cite{zhang2018unreasonable} on our synthetic dataset, along with non-reference metrics including FID \cite{Seitzer2020FID}, Inception Score \cite{salimans2016improvedtechniquestraininggans}, and MUSIQ \cite{ke2021musiq} on the in-the-wild dataset. For FID computation, we use the rendered images from Objaverse to represent the real distribution. 

\begin{table}[t]
\centering
\caption{Quantitative comparisons of enhancing multi-view synthesis on the Objaverse synthetic dataset, the best and second-best results are marked in \red{\underline{red}} and \blue{blue}, respectively.}
\vspace{-2mm}
  \renewcommand{\arraystretch}{1.15}
  \renewcommand{\tabcolsep}{3mm}
  \scalebox{0.88}{
    \begin{tabular}{@{}lcccc@{}}
      \toprule
      Method & PSNR$\uparrow$ & SSIM$\uparrow$ & LPIPS $\downarrow$ \\
      \midrule
      Input & 26.15 & 0.9056 & 0.1257 \\
      Real-ESRGAN~\cite{wang2021realesrgan} & 26.02 & 0.9185 & \blue{0.0877} \\
      StableSR~\cite{wang2024stablesr} & 25.12 & 0.8914 & 0.1130 \\
      RealBasicVSR~\cite{chan2022realbasicvsr} & \blue{26.21} & \blue{0.9212} & 0.0888 \\
      Upscale-A-Video~\cite{zhou2024upscale} & 25.57 & 0.8937 & 0.1153 \\
      3DEnhancer (Ours) & \red{\underline{27.53}} & \red{\underline{0.9265}} & \red{\underline{0.0626}} \\
      \bottomrule
    \end{tabular}
  }
\label{tab:synthetic_result}
\vspace{-1mm}
\end{table}

\subsection{Comparisons}

\begin{table}[t]
\centering
\caption{Quantitative comparisons of enhancing multi-view synthesis on the in-the-wild dataset.}
\vspace{-2mm}
  \renewcommand{\arraystretch}{1.15}
  \renewcommand{\tabcolsep}{3mm}
  \scalebox{0.85}{
    \begin{tabular}{@{}lcccc@{}}
      \toprule
      Method & MUSIQ$\uparrow$ & FID$\downarrow$ & IS$\uparrow$  \\
      \midrule
      Generated MV &52.77 &112.12 &7.68 $\pm$
    0.86 \\
      Real-ESRGAN~\cite{wang2021realesrgan} &72.47 &114.25 &7.31 $\pm$ 0.89\\
      StableSR~\cite{wang2024stablesr}   &70.43 &\blue{111.53} & 7.59 $\pm$ 0.97 \\
      RealBasicVSR~\cite{chan2022realbasicvsr} &\red{\underline{74.07}} &128.30 &7.09 $\pm$ 0.87 \\
      Upscale-A-Video~\cite{zhou2024upscale} &71.73 &114.81 &\blue{7.75 $\pm$ 0.96} \\
      3DEnhancer (Ours) & \blue{73.32} & \red{\underline{108.40}} & \red{\underline{7.93 $\pm$ 1.11}} \\
      \bottomrule
    \end{tabular}
  }
  \vspace{-6mm}
  \label{tab:real_result}
\end{table}

\noindent\textbf{Enhancing Multi-view Synthesis.}
The output images from multi-view synthesis models often lack texture details or exhibit inconsistencies across views, as shown in Fig.~\ref{fig:teaser}.
To demonstrate that \nickname{} can correct flawed textures and recover missing textures,
we provide quantitative results on both the Objaverse synthetic dataset and the in-the-wild dataset in Tab.~\ref{tab:synthetic_result} and Tab.~\ref{tab:real_result}, respectively. 
Qualitative comparisons on both test sets are presented in Fig.~\ref{fig:qualitative_synthetic} and Fig.~\ref{fig:qualitive_inthewild}. 
As can be seen,
our method outperforms others across most metrics. While RealBasicVSR achieves a higher MUSIQ score on the in-the-wild dataset, it fails to generate visually plausible images, as shown in Fig.~\ref{fig:qualitive_inthewild}. The image enhancement models RealESRGAN and StableSR can recover textures to some extent in individual views, but they fail to maintain consistency across multiple views. Video enhancement models, such as RealBasicVSR and Upscale-A-Video, also fail to correct texture distortions effectively. For example, both models fail to generate smooth facial textures in the first example shown in Fig.~\ref{fig:qualitive_inthewild}. In contrast, our method generates more natural and consistent details across views.

\begin{table}[ht]
\centering
\caption{Quantitative comparisons of enhancing 3D reconstruction on the in-the-wild dataset.}
\vspace{-2mm}
  \renewcommand{\arraystretch}{1.15}
  \renewcommand{\tabcolsep}{3mm}
  \scalebox{0.85}{
    \begin{tabular}{@{}lcccc@{}}
      \toprule
      Method & MUSIQ$\uparrow$ & FID$\downarrow$ & IS$\uparrow$  \\
      \midrule
      Input &41.04 &77.54 &8.98 $\pm$
    0.65 \\
      Real-ESRGAN~\cite{wang2021realesrgan} &65.25 &74.29 &8.29 $\pm$
    0.35 \\
      StableSR~\cite{wang2024stablesr}   &\blue{65.71} &\blue{72.98} &9.51 $\pm$
    0.78 \\
      RealBasicVSR~\cite{chan2022realbasicvsr} &65.70 &74.58 &7.77 $\pm$
    0.48 \\
      Upscale-A-Video~\cite{zhou2024upscale} &64.05 &74.12 &\blue{9.77 $\pm$
    0.79} \\
      3DEnhancer (Ours) & \red{\underline{66.04}} & \red{\underline{71.78}} & \red{\underline{9.96 $\pm$ 0.96}} \\
      \bottomrule
    \end{tabular}
  }
  \label{tab:3d_real_result}
\vspace{-6mm}
\end{table}

\begin{figure*}[ht]
\begin{center}\includegraphics[width=\linewidth]{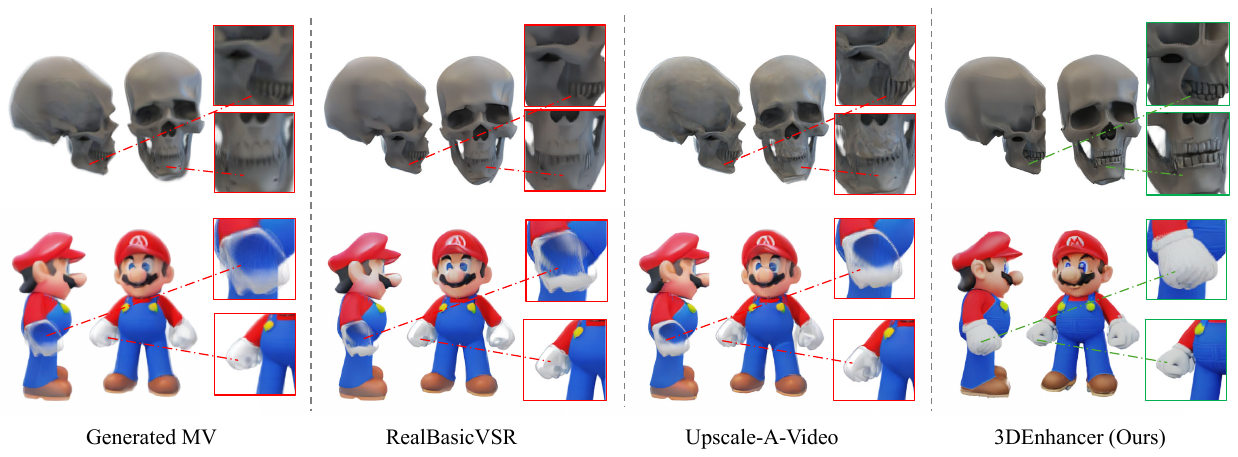}
    \vspace{-6mm}
    \caption{
    Qualitative comparisons of enhancing 3D reconstruction given generated multi-view images on the in-the-wild dataset. Multi-view models produce low-quality, view-inconsistent outputs, leading to flawed 3D reconstructions. Existing methods fail to correct texture artifacts, while our method produces both geometrically accurate and visually appealing results.
    }
    \vspace{-6mm}
\label{fig:qualitive_inthewild_3d}
\end{center}
\end{figure*}

\begin{figure}[t]
\begin{center}
    \includegraphics[width=.99\linewidth]{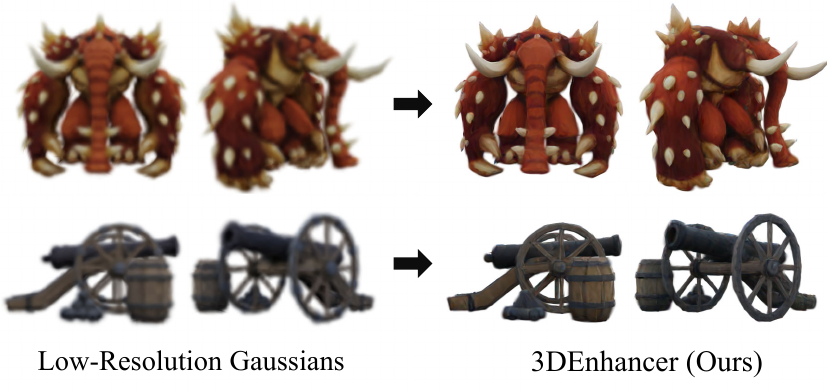}
    \vspace{-3mm}
    \caption{
    Low-resolution GS optimization with \nickname{}.
    }
    \vspace{-4mm}
\label{fig:optimize_gaussian}
\end{center}
\end{figure}

\noindent\textbf{Enhancing 3D Reconstruction.} In this section, we present 3D reconstruction comparisons based on rendering views from 3DGS generated by LGM \cite{tang2024lgm}. Quantitative comparisons are shown in Tab.~\ref{tab:3d_real_result}. Our method outperforms previous approaches in terms of 3D reconstruction. For qualitative evaluation, we visualize the results of two video enhancement models, RealBasicVSR and Upscale-A-Video. As shown in Fig.~\ref{fig:qualitive_inthewild_3d}, these baselines suffer from a lack of multi-view consistency, leading to misalignment, such as the misalignment of the teeth in the first skull example and the ghosting in the example of Mario's hand. In contrast, our model maintains consistency and produces high-quality texture details. We further demonstrate our approach can optimize coarse differentiable representations. As shown in Fig.~\ref{fig:optimize_gaussian}, our method is capable of refining low-resolution Gaussians~\cite{Shen2024SuperGaussian}. More details and results of refining coarse Gaussians are provided in the Appendix Sec.~\ref{sec:optimization}.

\begin{table}[h]
\centering
\vspace{-2.6mm}
\caption{Ablation study of cross-view modules.}
\vspace{-2mm}
\renewcommand{\arraystretch}{1.15}
\renewcommand{\tabcolsep}{1.5mm}
\scalebox{0.82}{
\begin{tabular}{c|cc|ccc}
\toprule
Exp. & Multi-view Attn. & Epipolar Agg. &    PSNR$\uparrow$ & SSIM$\uparrow$ & LPIPS $ \downarrow$\\ 
\midrule
(a)   &   & & 25.11 & 0.9067 & 0.081\\
(b)  &  & \Checkmark  &25.95 & 0.9147 & 0.072 \\ 
(c) &\Checkmark&  &26.92 & 0.9226    &  0.0642   \\ 
(d) &\Checkmark & \Checkmark  &  {\bf 27.53}   &  {\bf 0.9265} & {\bf 0.0626}   \\ \bottomrule
\end{tabular}
}
\label{tab:ablation}
\vspace{-4mm}
\end{table}

\subsection{Ablation Study}

\begin{figure}[t]
\begin{center}
    \includegraphics[width=\linewidth]{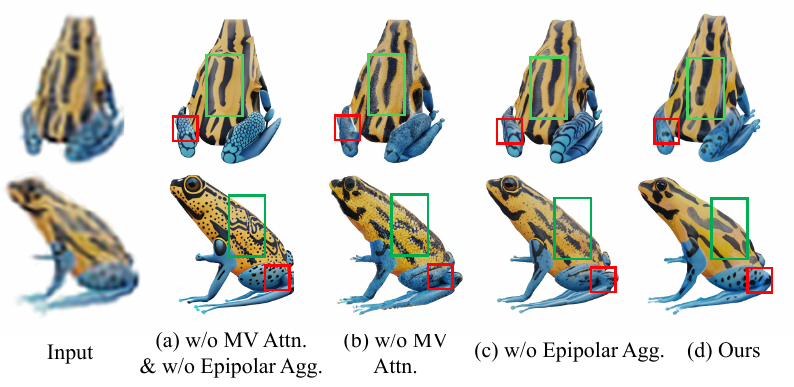}
    \vspace{-6mm}
    \caption{
    Effectiveness of cross-view modules.
    }
    \vspace{-6mm}
\label{fig:abalation_all}
\vspace{-2mm}
\end{center}
\end{figure}

\noindent\textbf{Effectiveness of Cross-View Modules.} To evaluate the effectiveness of our proposed cross-view modules, we ablate two modules: multi-view row attention and near-view epipolar aggregation. As shown in Tab.~\ref{tab:ablation}, removing either module results in worse textures between views. The visual comparison in Fig.~\ref{fig:abalation_all} also validates this observation.
Without the multi-view row attention module, the model fails to produce smooth textures, as shown in Fig.~\ref{fig:abalation_all}~(b).  
Without the epipolar aggregation module, reduced texture consistency is observed, as depicted in Fig.~\ref{fig:abalation_all}~(c).

\begin{figure}[t]
\begin{center}  \includegraphics[width=\linewidth]{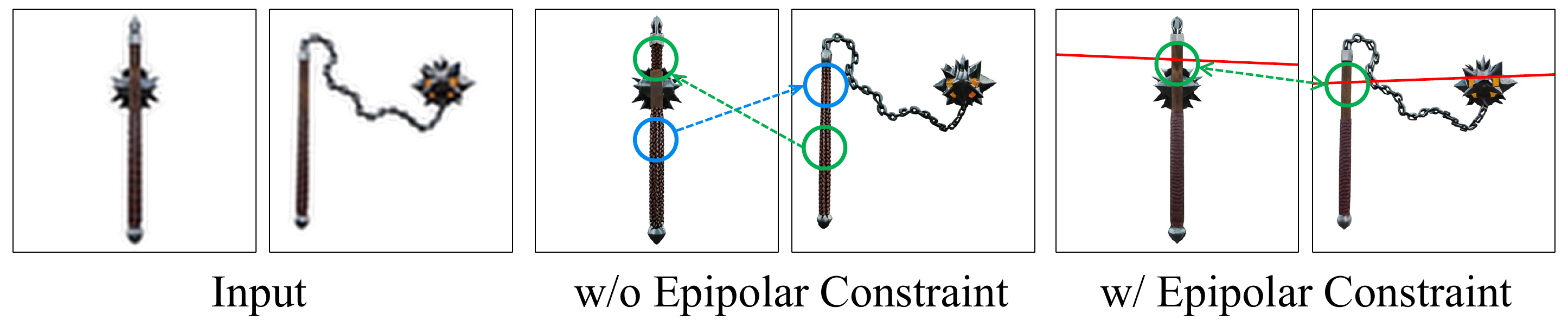}
    \vspace{-4mm}
    \caption{
    Comparisons of enhancing multi-view images with and without epipolar aggregation. The red line denotes the epipolar line corresponding to the circled area, while the dotted arrow indicates the corresponding area from one view to another.
    }
    \vspace{-8mm}
\label{fig:epipolar}
\end{center}
\end{figure}

Besides, the epipolar constraint is essential for preventing the model from learning textures from incorrect regions in other views and contributes to the overall consistency. As demonstrated in Fig.~\ref{fig:epipolar}, without the epipolar constraint, the texture of the top part of the flail is incorrectly aggregated from the grip in the other view, thus resulting in inconsistency across views.

\vspace{2mm}

\noindent\textbf{Effectiveness of Noise Level.} As shown in Fig.~\ref{fig:teaser}, our model can generate diverse textures by adjusting noise levels. Low noise levels generally result in outputs with blurred details, while high noise levels produce sharper, more detailed textures. However, high noise levels may also reduce the fidelity of the input images.
\section{Conclusion}
\label{sec:conclusion}
In conclusion, this work presents a novel 3D enhancement framework that leverages view-consistent latent diffusion model to improve the quality of given coarse multi-view images. Our approach introduces a versatile pipeline combining data augmentation, multi-view attention and epipolar aggregation modules that effectively enforces view consistency and refines textures across multi-view inputs. Extensive experiments and ablation studies demonstrate the superior performance of our method in achieving high-quality, consistent 3D content, significantly outperforming existing alternatives. This framework establishes a flexible and powerful solution for generic 3D enhancement, with broad applications in 3D content generation and editing.

\vspace{2mm}
\noindent{\bf Acknowledgement.} This study is supported under the RIE2020 Industry Alignment Fund – Industry Collaboration Projects (IAF-ICP) Funding Initiative, as well as cash and in-kind contribution from the industry partner(s).  It is also supported by Singapore MOE AcRF Tier 2 (MOE-T2EP20221-0011) and the National Research Foundation, Singapore, under its NRF Fellowship Award (NRF-NRFF16-2024-0003).

{
    \small
    \bibliographystyle{ieeenat_fullname}
    \bibliography{main}
}

\clearpage
\renewcommand\thesection{\Alph{section}}
\onecolumn
\setcounter{section}{0}


%
\begin{center}
	\Large\textbf{{Appendix}}\\
	\vspace{8mm}
\end{center}

\maketitle
In this appendix, we provide additional discussions and results to supplement the main paper. 
In Sec.~\ref{sec:arch}, we present more architecture and design details of our \NICKNAME{}.
In Sec.~\ref{sec:dataset}, we provide detailed information about our training dataset, including the augmentation pipeline and illustrative examples. Sec.~\ref{sec:inference_details} highlights some interesting findings related to inference. More results and comparisons are presented in Sec.~\ref{sec:results} to further demonstrate our performance. We also include a demo video (Sec.~\ref{subsec:demo_video}) to showcase rendering results for 3D reconstruction enhancement.

{   
    \hypersetup{linkcolor=blue}
    \tableofcontents
}

\clearpage

\section{Architecture and Design}
\label{sec:arch}

\subsection{Pose-aware Encoder}
Our pose-aware encoder is adapted from the convolutional encoder of LDM~\cite{rombach2022LDM}. As shown in Fig.~\ref{fig:overview}, the output of the pose-aware encoder serves as the conditioning features for the trainable copies in our ControlNet~\cite{zhang2023adding}. The details of its hyperparameters are summarized in Tab.~\ref{tab:arch}. This encoder employs 64 channels and a single residual block to enhance efficiency. Additionally, we incorporate cross-view self-attention~\cite{shi2023MVDream} into the middle layer of the encoder to improve inter-view consistency. To ensure compatibility with the number of latent channels in the DiT blocks, the output $z$-channels number is set to 1152. The final convolutional layer in the encoder uses a stride of 2 to match the dimensions of the DiT block latents. All other hyperparameters are kept at default values.

\subsection{View-Consistent DiT Block}
The view-consistent DiT block is based on the PixArt-$\Sigma$~\cite{chen2024pixartsigma} architecture. Consistent with PixArt-$\Sigma$, we use the T5 large language model as the text encoder for conditional text feature extraction, and the frozen VAE from SDXL~\cite{sdxl} to capture the latent features of images. PixArt-$\Sigma$ consists of 28 Transformer blocks. For the ControlNet~\cite{zhang2023adding} implementation, we utilize trainable copies of the first 13 base blocks, augmenting each copied block with zero linear layers before and after it. The output of the $i$-th trainable copied block is added to the corresponding frozen base $i$-th block. The multi-view row attention with near-view epipolar aggregation is an additional attention layer that is inserted into both the DiT blocks and the copied ControlNet blocks. This layer is positioned after the self-attention layer, as illustrated in Fig.~\ref{fig:overview}. During training, we train the entire ControlNet blocks and every inserted multi-view row attention layer in the DiT blocks. Detailed hyperparameters for the DiT block and the inserted row attention layers are provided in Tab.~\ref{tab:arch}.

\subsection{Weight for Two Nearest Views Aggregation}
\label{sec:weight}
In Eq.~\ref{eq:aggregate_view}, we compute the fusion weight \( w \) based on both the physical camera distance and the similarity of token features. First, we consider the geometric distance weight \( w_d \), which reflects the proximity of the camera:

\begin{equation}
w_d = \frac{d_{\rvv, \rvv+1}}{d_{\rvv, \rvv-1} + d_{\rvv, \rvv+1}},
\end{equation}
where \( d_{\rvv, \rvk} \) represents the geometric distance between the camera of view \( v \) and the camera of view $\rvk \in \{\rvv-1, \rvv+1\}$. To ensure the nearest-view weight calculation also incorporates token feature similarity, we augment the weight token-wise with token similarity:

\begin{equation}
w = \frac{S^i_{\rvv, \rvv-1} \cdot w_d}{S^i_{\rvv, \rvv-1} \cdot w_d + (1 - w_d) \cdot S^i_{\rvv, \rvv+1}},
\end{equation}
where \( S^i_{\rvv, \rvk} \) denotes the cosine similarity of the corresponding tokens, \ie, $\rvf_\rvv[i]$ and $\rvf_{\rvk}[M_{\rvv, \rvk}[i]]$.

\begin{table}[h]
\caption{Hyperparameters for the pose-aware encoder, view-consistent DiT block, and the inserted multi-view row attention layers in our \NICKNAME{}. The table follows the hyperparameter table style from \cite{rombach2022LDM, Peebles2022DiT}. We train our model on images with a resolution of $512 \times 512$ using 4 views.}

\centering
\begin{minipage}[t]{0.36\textwidth}
\centering
\renewcommand{\arraystretch}{1}
\renewcommand{\tabcolsep}{2.05mm}
\resizebox{\linewidth}{!} {
\begin{tabular}{lc}
\toprule
Hyperparameter &  \textit{DiT} \\ 
\midrule
Layers & 28 \\
Training views shape   &  $4\times 512\times 512 \times 3$ \\
$f$  & 8 \\
Patch size &  2 \\
Embedding dimension & 1024 \\
Hidden size &  1152 \\
$z$-shape   &  $4\times 1024\times  1152$ \\
Head number & 16 \\
CA sequence length & 300 \\
\bottomrule
\end{tabular}
}
\label{tab:arch}
\end{minipage}
\hspace{8mm}
\begin{minipage}[t]{0.33\textwidth}
\centering
\renewcommand{\arraystretch}{1}
\renewcommand{\tabcolsep}{2.05mm}
\resizebox{\linewidth}{!} {
\begin{tabular}{lc}
\toprule
Hyperparameter &  \textit{Pose-aware Encoder} \\
\midrule
$f$  & 8 \\
Channels & 64 \\
Channel multiplier & 1, 2, 4, 4 \\
$z$-channels & 1152 \\
\bottomrule
\vspace{1.5mm}\\
\toprule
Hyperparameter &  \textit{Row Attention} \\
\midrule
Head number & 16 \\
Positional encoding & sine-cosine \\
Epipolar aggregation & True \\
\bottomrule
\end{tabular}
}
\label{tab:arch}
\end{minipage}

\end{table}

\section{Dataset}
\label{sec:dataset}

\subsection{Dataset}

The G-buffer Objaverse dataset~\cite{qiu2024richdreamer} contains a broad variety of 3D objects categorized into 10 types: Human-Shaped, Animals, Daily Objects, Furniture, Buildings and Outdoor Objects, Transportation, Plants, Food, and Electronics. To ensure high standards, we exclude any objects labeled as “Poor-quality.” We observe that the original captions in G-buffer Objaverse are simple and lack detailed information. Therefore, we adopt captions from 3D-Topia~\cite{hong20243dtopia}, which provide more informative and accurate descriptions for a subset of objects in Objaverse. We update the caption of each object accordingly if it exists in 3D-Topia, resulting in the refinement of approximately 45\% of the captions. Additionally, to facilitate CFG~\cite{Ho2022ClassifierFreeDG}, we omit the text condition at a rate of 0.2. Such settings enhance the robustness of our method to text conditions with varying levels of detail. For the in-the-wild dataset, we remove backgrounds and center objects as previous works~\cite{wang2023imagedream, liu2023syncdreamer, li2024era3d}. We uniformly apply a white background to the input views.

\begin{figure*}[h]
\begin{center}
\includegraphics[width=.95\linewidth]{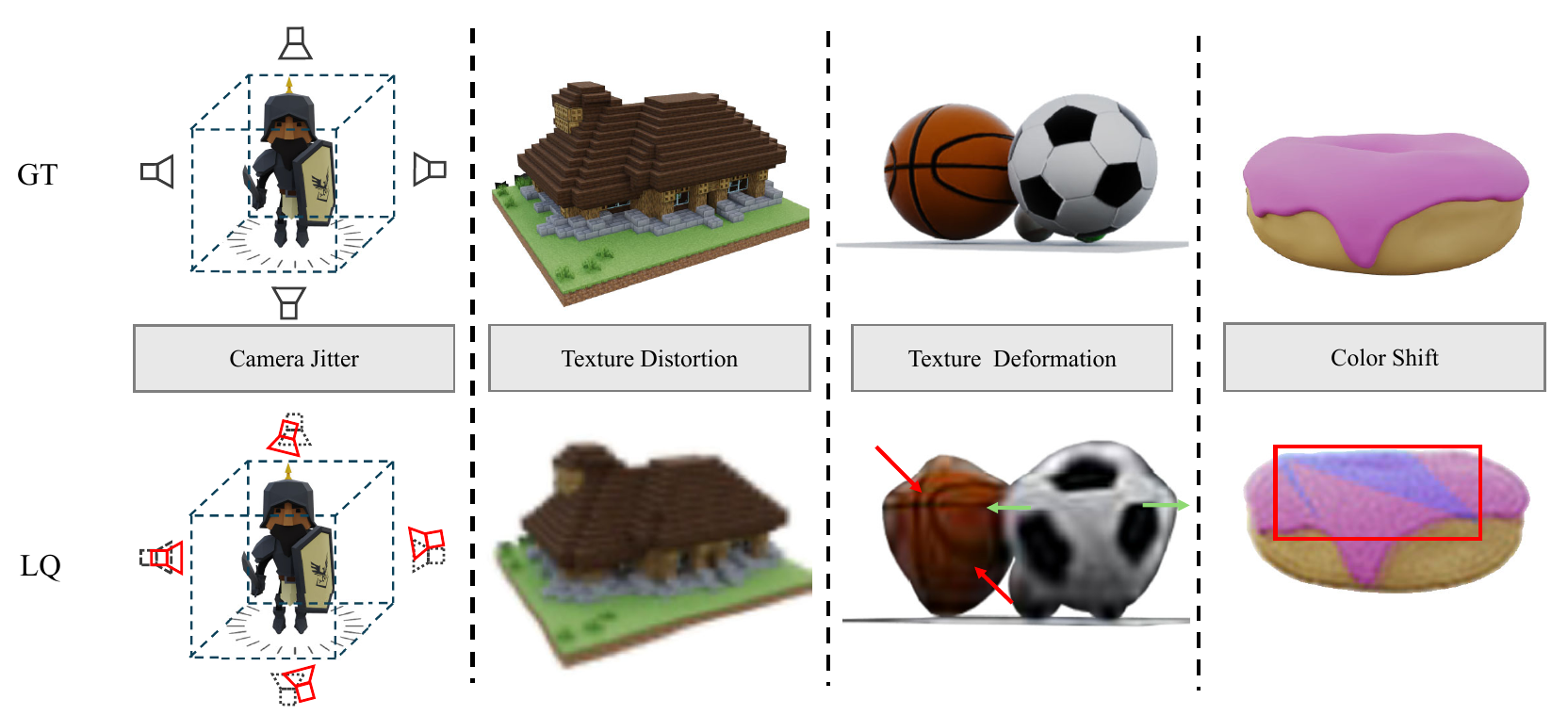}
    \vspace{-1mm}
    \caption{
        Visualization of several examples from our augmentation pipeline. Thanks to the comprehensive augmentation strategy, our method is able to bridge the domain gap between training and inference.
    }
    \vspace{-6mm}
\label{fig:augmentation_pipeline}
\end{center}
\end{figure*}

\subsection{Data Augmentation}
The visualization of the data augmentation pipeline is shown in Fig.~\ref{fig:augmentation_pipeline}. During training, we dynamically generate synthetic training pairs on the fly, and the argumentation is implemented in PyTorch with CUDA acceleration to ensure efficiency. The pipeline incorporates several stochastic augmentation steps, producing diverse training pairs with varying levels of degradation. During augmentation, the input views of the same object are either augmented with the same level of degradation (e.g., the same blur kernel) or with different stochastic augmentations. This strategy encourages the model's ability to learn information across views, particularly from those with fewer degradations. We ensure that the augmentation is confined to the object's masked area with a slight mask dilation. This allows the white background unaffected, which aligns with real-world scenarios of low-quality multi-view images. We also set a probability where no augmentation is applied to the input images, i.e., the low-quality images are identical to the ground truth. In such cases, the model is encouraged to preserve fidelity when the input images are already of high-quality. Details of several augmentation parameters are summarized in Tab.~\ref{tab:augmentation_params}. Further implementation details will be provided in our code release.

\begin{table}[h]
\caption{Several augmentation parameters that are used in our augmentation pipeline.}

\centering
\begin{minipage}[t]{0.4\textwidth}
\centering
\renewcommand{\arraystretch}{1}
\renewcommand{\tabcolsep}{2.05mm}
\resizebox{\linewidth}{!} {
\begin{tabular}{lc}
\toprule
Argumentation type &  Parameters \\ 
\midrule
First-order blur prob & 0.8 \\
Second-order blur prob & 0.3 \\
Blur kernel size range & \{7, 9, ..., 21\} \\
Blur standard deviation range   &  [0.2, 3] \\
Gaussian noises prob & 0.5 \\
Resize range & [0.3, 1.5] \\
JPEG compression quality factor & [80, 100] \\
\bottomrule
\end{tabular}
}
\label{tab:augmentation_params}
\end{minipage}
\hspace{8mm}
\begin{minipage}[t]{0.36\textwidth}
\centering
\renewcommand{\arraystretch}{1}
\renewcommand{\tabcolsep}{2.05mm}
\resizebox{\linewidth}{!} {
\begin{tabular}{lc}
\toprule
Argumentation type &  Parameters \\ 
\midrule
Final sinc filter prob  & 0.8 \\
Camera jitter prob & 0.2 \\
Camera jitter strength range & [0.05, 0.1] \\
Color shift prob & 0.3 \\
Grid distortion prob & 0.3 \\
Grid distortion strength range &  [0.2, 0.5] \\
No argumentation prob  & 0.1 \\
\bottomrule
\end{tabular}
}
\label{tab:augmentation_params}
\end{minipage}

\end{table}

\section{More Details on Inference}
\label{sec:inference_details}
\subsection{Multi-View Editing}
\begin{figure}[h]
\begin{center}
    \vspace{-3mm}
    \includegraphics[width=.90\linewidth]{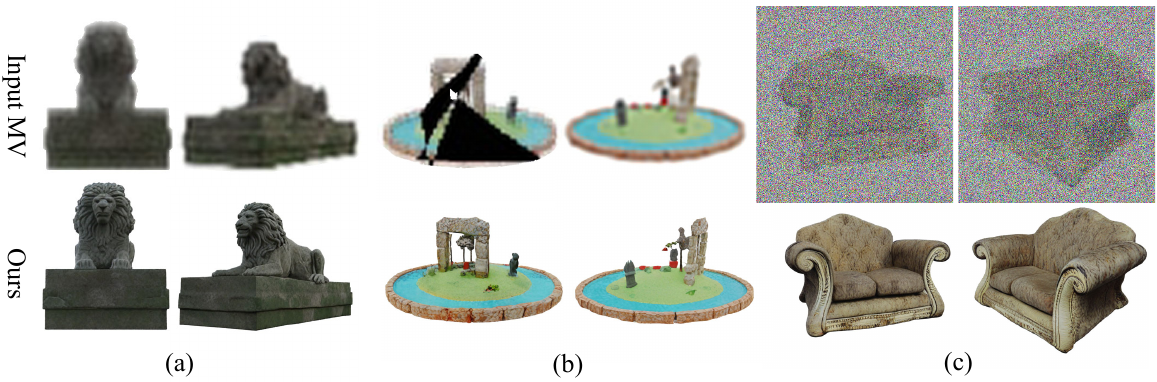}
    \caption{
        Examples of handling extremely coarse inputs with \NICKNAME{}.
    }
    \vspace{-4mm}
\label{fig:rebuttal_coarse_results}
\end{center}
\end{figure}

Benefiting from our comprehensive augmentation pipeline and the robust view-consistent DiT Block, we observe an interesting fact: our method is capable of generating detailed and consistent textures even from extremely coarse or corrupted multi-view inputs. As shown in Fig.~\ref{fig:rebuttal_coarse_results}, our method effectively handles various challenging cases, including multi-views with (a) \textit{extremely blurred textures}, (b) \textit{masked or missing parts}, and (c) \textit{significant noise}. 

\begin{figure*}[h]
\begin{center}
\includegraphics[width=.95\linewidth]{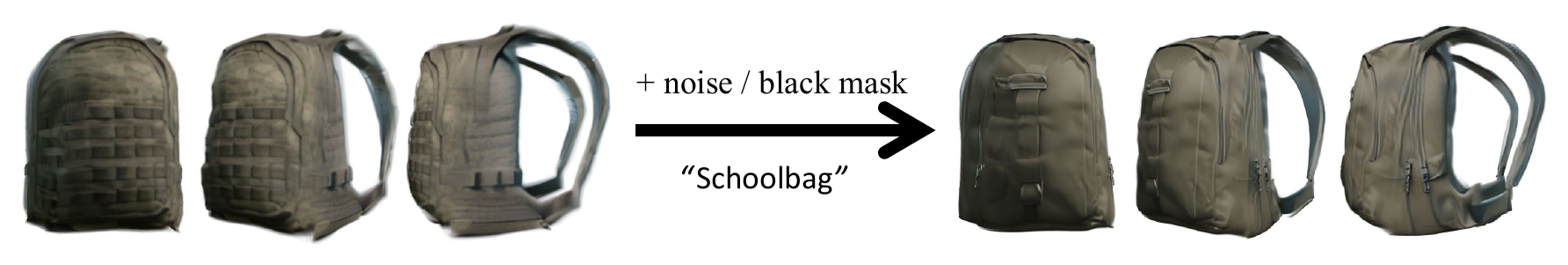}
    \vspace{-1mm}
    \caption{
        Rendered views of edited 3D Gaussians using our multi-view editing approach. By adding a large noise or a black mask, and leveraging text prompts as guidance, we consistently modify the texture of the bags.
    }
    \vspace{-6mm}
\label{fig:mv_edit}
\end{center}
\end{figure*}

\noindent This enables our approach to modify multi-view images in two distinct ways: 1. Applying a black mask to the region designated for editing and modifying the text prompt to generate the target multi-view images.  
2. Adjusting the inference noise level, where higher noise levels produce more diverse outputs.  Using the edited multi-view images, we can subsequently modify the reconstructed 3D representations. An example of editing 3D Gaussians generated by LGM \cite{tang2024lgm} through modifying its multi-view input is shown in Fig.~\ref{fig:mv_edit}.

\subsection{Color Correction}
\begin{figure*}[h]
\begin{center}
\includegraphics[width=.9\linewidth]{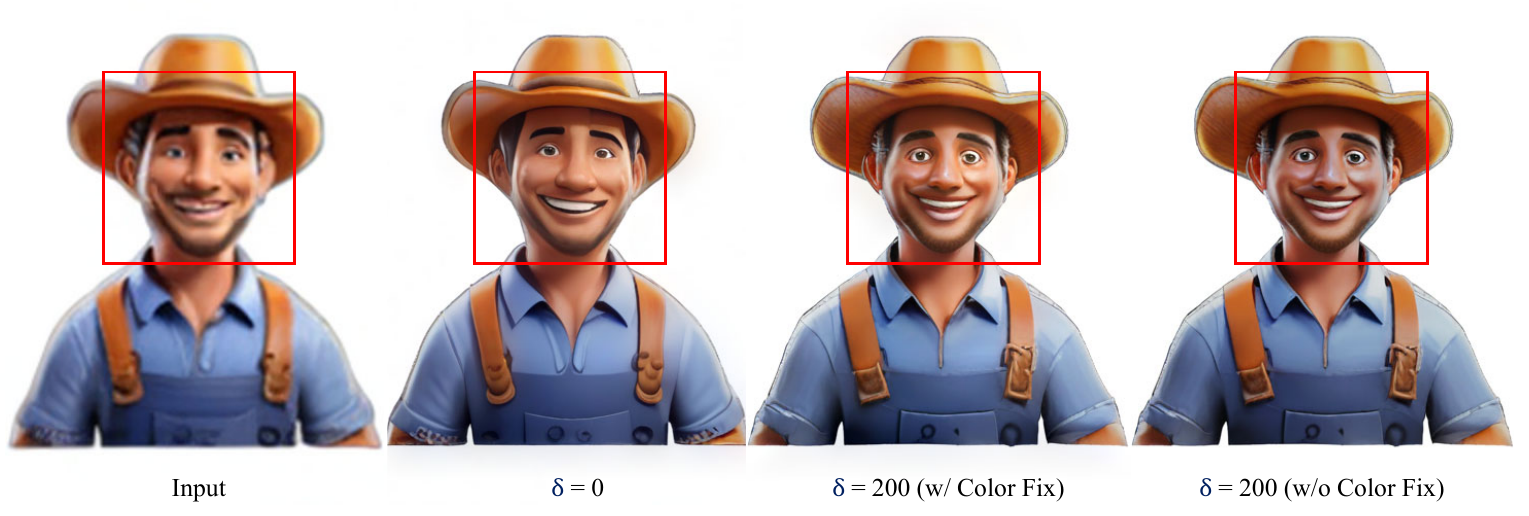}
    \vspace{-2mm}
    \caption{
        Minor global color scheme shift at high noise levels. When the noise level $\delta$ is small, such as $\delta = 0$, our method maintains excellent color fidelity. However, at a higher noise level, such as $\delta = 200$ in the example, the output figure's face appears slightly darker than that of the input. In this case, the wavelet color correction~\cite{wang2024stablesr} could help mitigate this issue.
    }
    \vspace{-6mm}
\label{fig:color_shift_supp}
\end{center}
\end{figure*}

Previous studies~\cite{wang2024stablesr, zhou2024upscale} have highlighted that diffusion models often exhibit color shift artifacts, where the global color scheme deviates from the input images. This is different from our color shift augmentation, which introduces localized color changes to specific image regions. However, this augmentation also aims to encourage the model to maintain consistent color reproduction. We observe that integrating a training-free wavelet color correction module~\cite{wang2024stablesr} can help resolve the global color scheme shift. As reported in Tab.~\ref{tab:unique3d}, applying wavelet color correction leads to improved fidelity metrics (higher PSNR, SSIM, and lower LPIPS~\cite{zhang2018unreasonable}) for the baseline, but it has minimal impact on our results, showing our robustness against global color scheme shifts. However, at extremely high noise levels, such as $\delta = 200$, minor global color shifts may still occur in our method because the noise may impact the original color information. In such cases, wavelet color correction could be beneficial, as illustrated in Fig.~\ref{fig:color_shift_supp}.

\section{More Results}
\label{sec:results}
\subsection{User Study}
\begin{figure*}[h]
\begin{center}
\includegraphics[width=.6\linewidth]{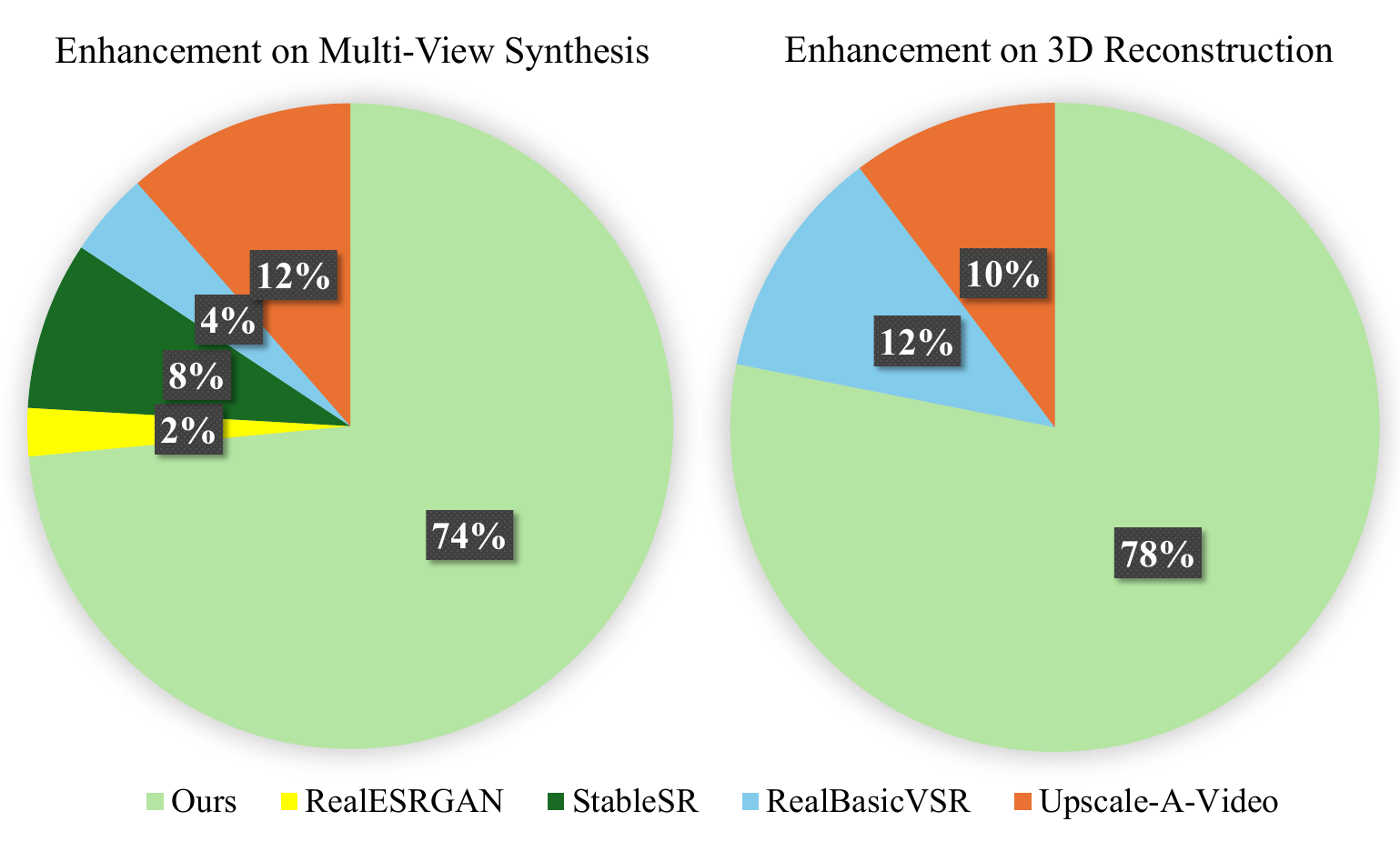}
    \vspace{-1mm}
    \caption{
        User study results. Human voters consistently prefer our method over other approaches.
    }
    \vspace{-4mm}
\label{fig:user_study}
\end{center}
\end{figure*}

To enable a thorough comparison, we conduct a user study to evaluate the enhancement results of multi-view images and 3D reconstructions. For the multi-view image enhancement, each participant is shown 10 sets of randomly selected objects' multi-view images, enhanced by our \NICKNAME{},  RealESRGAN~\cite{wang2021realesrgan}, StableSR~\cite{wang2024stablesr}, RealBasicVSR~\cite{chan2022realbasicvsr}, and Upscale-a-Video~\cite{zhou2024upscale}. For the 3D reconstruction enhancement, participants are presented with another 10 360-degree rotating render videos of the 3D Gaussians enhanced by our method, RealBasicVSR~\cite{chan2022realbasicvsr}, and Upscale-a-Video~\cite{zhou2024upscale}. Their task is to choose the visually superior enhanced results. A total of 20 participants take part in the study. As illustrated in Fig.~\ref{fig:user_study}, The results indicate a strong preference for our method over the compared approach. On average, 74\% of users preferred our method for enhancing multi-view images, while 78\% favored it for enhancing 3D reconstruction. These findings strongly demonstrate the quality and robustness of our approach.

\subsection{Results of Optimizing 3D Gaussians}
\label{sec:optimization}
\begin{figure*}[h]
\begin{center}
\includegraphics[width=\linewidth]{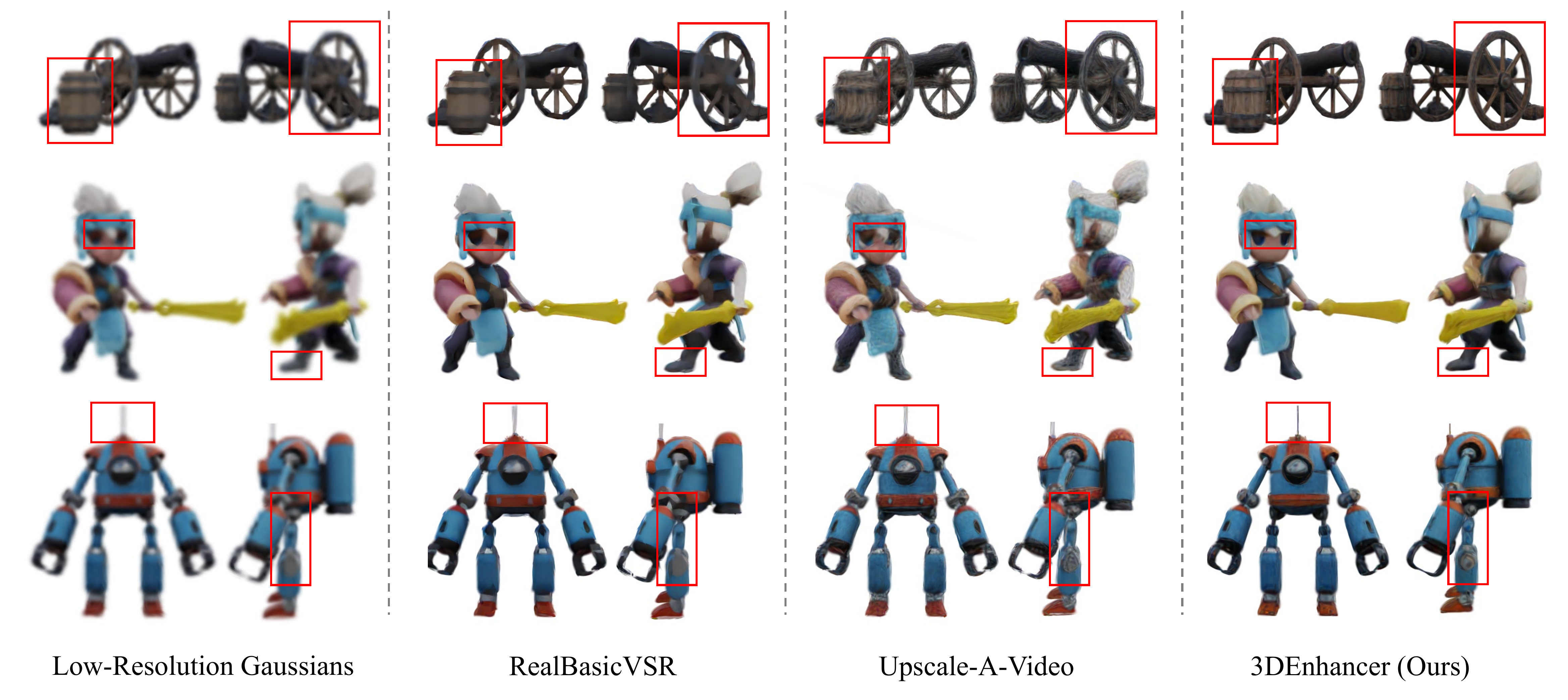}
    \vspace{-4mm}
    \caption{
       Qualitative comparisons of optimizing low-resolution Gaussians. During optimization, both RealBasicVSR~\cite{chan2022realbasicvsr} and Upscale-A-Video ~\cite{zhou2024upscale} produce ghosting and blurry textures due to inconsistent outputs. Our \NICKNAME{} achieves sharp and clear results.
    }
    \vspace{-4mm}
\label{fig:3d_optimize_supp}
\end{center}
\end{figure*}

3D representations can be rendered from multiple views, this nature allows our method to iteratively optimize a coarse 3D representations. To demonstrate this capability, we adopt Gaussian Splatting \cite{kerbl3Dgaussians} as our example due to its high rendering fidelity and efficiency. Specifically, we implement a pipeline to refine coarse 3D Gaussians checkpoints by leveraging our enhanced outputs as pseudo ground truth. We randomly select 20 objects from the Objaverse test dataset for evaluation. Following \cite{Shen2024SuperGaussian}, we fit low-resolution 3D Gaussians using images obtained by bilinearly downsampling the original dataset images by a factor of 8, resulting in a resolution of 64 × 64 pixels. We use three distinct trajectories for fitting low-resolution Gaussians, refining Gaussians, and evaluation. As proposed in \cite{gao2024cat3d}, our refinement process also minimizes a combined loss function, including a photometric reconstruction loss and a perceptual loss~\cite{zhang2018unreasonable}. The perceptual loss emphasizes high-level semantic similarity between rendered and enhanced images while ignoring inconsistencies in low-level, high-frequency details. To improve regularization during refining, we sample 100 views along a single smooth orbital path, as increasing the number of views has been shown to enhance the refining process~\cite{gao2024cat3d}. The optimization is conducted over 2000 refinement steps for all methods and takes approximately 130s to refine a single object on one NVIDIA A100 GPU. For comparison, we evaluate our method against two video enhancement models, RealBasicVSR ~\cite{chan2022realbasicvsr} and Upscale-A-Video~\cite{zhou2024upscale}. Quantitative and qualitative results are presented in Tab.~\ref{tab:3d_optimize_supp} and Fig.~\ref{fig:3d_optimize_supp}, respectively. Our results demonstrate detailed and sharp outputs, while other methods exhibit ghosting artifacts and blurry textures. The results highlight the superior performance of our approach in refining coarse 3D representations.

\begin{table}[h]
\caption{Quantitative comparisons of optimizing low-resolution Gaussians. The best results are highlighted in \textbf{bold}.}
\centering
\vspace{-1mm}
\renewcommand{\arraystretch}{1.15}
\renewcommand{\tabcolsep}{2.05mm}
\resizebox{0.8\linewidth}{!} {
\begin{tabular}{c|cccc}
\toprule
Metrics & Low-Resolution Gaussians & RealBasicVSR \cite{chan2022realbasicvsr} &  Upscale-A-Video\cite{zhou2024upscale}  & \NICKNAME{} \\ 
\midrule
 PSNR $\uparrow$   &  26.35  & 27.39 & 26.20 & \textbf{27.54} \\
 SSIM $\uparrow$  & 0.9120  & 0.9216  & 0.9184   &  \textbf{0.9337} \\ 
 LPIPS $\downarrow$  & 0.1135  & 0.0803  & 0.0928   &  \textbf{0.0756} \\ 
\bottomrule
\end{tabular}
}
\label{tab:3d_optimize_supp}
\vspace{-3mm}
\end{table}

\subsection{Results of Generalization to Real-World Objects}  
\begin{figure}[h]
\begin{center}
    \vspace{-2mm}
    \includegraphics[width=.90\linewidth]{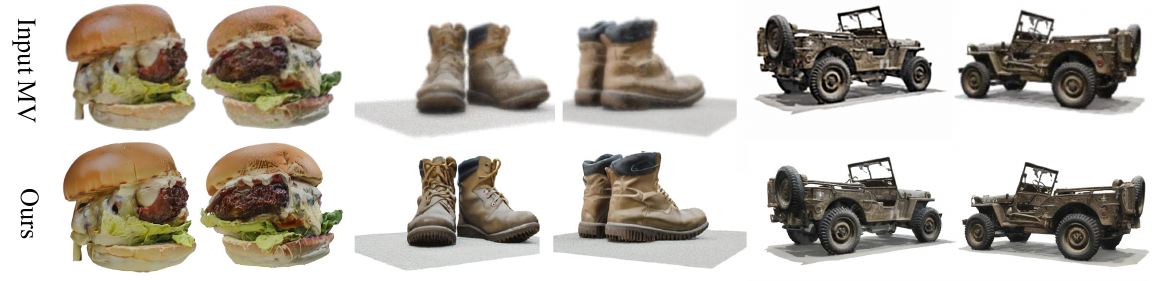}
    \vspace{1mm}
    \caption{
        Examples of handling complex real-world objects with \NICKNAME{}. Our method generates rich textures on realistic objects.
    }
    \vspace{-6mm}
\label{fig:rebuttal_real_test_results}
\end{center}
\end{figure}

We test our model on the constructed OmniObject3D dataset~\cite{wu2023omniobject3d}, which provides realistic 3D object scans, and also on complex, richly textured objects from Polycam. Backgrounds are removed as needed using BiRefNet~\cite{zheng2024birefnet}. As shown in Fig.~\ref{fig:rebuttal_real_test_results}, our model effectively enhances real-world objects.  

\subsection{Results of Further Fine-tuning Upscale-A-Video}
\begin{figure}[h]
\begin{center}
    \includegraphics[width=.90\linewidth]{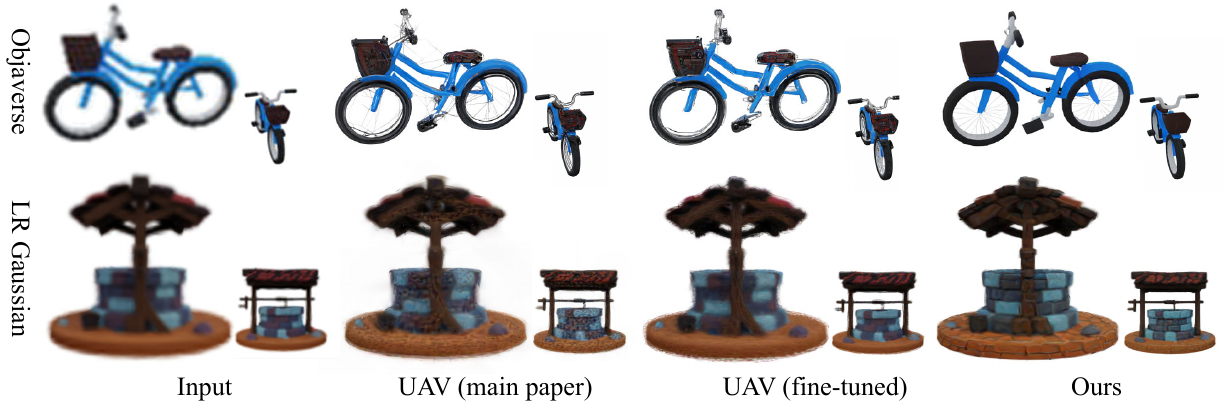}
    \vspace{1mm}
    \caption{  
        Qualitative comparisons with fine-tuned Upscale-A-Video (UAV) on synthetic Objaverse multi-view images and Low-Resolution (LR) Gaussians. With additional fine-tuning using our augmentations, Upscale-A-Video reduces inconsistent artifacts outside the object area. Our method still shows superior generative capabilities.  
    }  
    \vspace{-6mm}
\label{fig:rebuttal_finetune_uav}
\end{center}
\end{figure}

Our work aims to provide a \textit{generic} framework for 3D object enhancement, supporting enhancing (\textbf{I}) sparse multi-view images from large angles for multi-view reconstruction networks (\eg, LGM~\cite{tang2024lgm}), and (\textbf{II}) coarse 3D model via per-instance optimization. Existing video diffusion models, \eg., Upscale-A-Video~\cite{zhou2024upscale}, mainly rely on temporal attention for consistency. They are designed for handling adjacent video frames with minimal spatial variations, without considering camera pose. Thus, they struggle to establish multi-view correspondences in case (I), where input views vary significantly, leading to suboptimal results. Additionally, due to the huge GPU memory cost of video diffusion models, they also cannot handle dense 360° views simultaneously, typically working with short video sequences (\eg, 8 frames for Upscale-A-Video), limiting their performance in case (II) as well. Thus, this study is crucial to explore new and effective modules of \textit{sparse} multi-view attention for 3D enhancement, using a pose-aware encoder and an epipolar aggregation mechanism, which together achieve superior results in both (I) and (II) (see Tabs.~{\color{cvprblue}1-3} and Figs.~{\color{cvprblue}3} and {\color{cvprblue}4}). All baseline methods in the main paper are fine-tuned on the Objaverse dataset. We further fine-tune Upscale-A-Video with our proposed data augmentation. The results in the Tab.~\ref{tab:rebuttal_finetune_uav} and Fig.~\ref{fig:rebuttal_finetune_uav} show that our method still outperforms the video-based Upscale-A-Video, further supporting our discussion here.

\begin{table}[h]
\caption{Quantitative comparisons with fine-tuned Upscale-A-Video (UAV) on synthetic Objaverse multi-view images and Low-Resolution (LR) Gaussians.}  
\centering
\renewcommand{\arraystretch}{1}
\renewcommand{\tabcolsep}{1.5mm}
\resizebox{0.85\linewidth}{!} {
\begin{tabular}{c|ccc|ccc}
\toprule
\multirow{2}{*}{Method} & \multicolumn{3}{c|}{\textit{Synthetic Objaverse}} & \multicolumn{3}{c}{\textit{Low-Resolution Gaussians}} \\
\cmidrule(r){2-4} \cmidrule(l){5-7}
        & PSNR$\uparrow$ & SSIM$\uparrow$  & LPIPS$\downarrow$ & PSNR$\uparrow$ & SSIM$\uparrow$  & LPIPS$\downarrow$ \\
\hline
\midrule
\rowcolor{gray!20} UAV (main paper) & 25.57&  0.8937&   0.1153 & 26.20 & 0.9184 & 0.0928 \\
\rowcolor{gray!20} UAV (further fine-tuned) & 26.14{\scriptsize(\textcolor{red}{+0.57})}&  0.9086{\scriptsize(\textcolor{red}{+0.0149})}&  0.0996{\scriptsize(\textcolor{red}{-0.0157})} & 26.69 {\scriptsize(\textcolor{red}{+0.49})}&  0.9197 {\scriptsize(\textcolor{red}{+0.0013})}&  0.0850 {\scriptsize(\textcolor{red}{-0.0078})}  \\ 
Ours & \textbf{27.53} & \textbf{0.9265} & {\bf 0.0626} &    \textbf{27.54} & \textbf{0.9337} & {\bf 0.0756} \\ 
\bottomrule
\end{tabular}
}
\label{tab:rebuttal_finetune_uav}
\end{table}

\newpage
\subsection{More Comparisons}
\label{subsec:more_comparisons}
\begin{figure*}[h]
\begin{center}
\includegraphics[width=.88\linewidth]{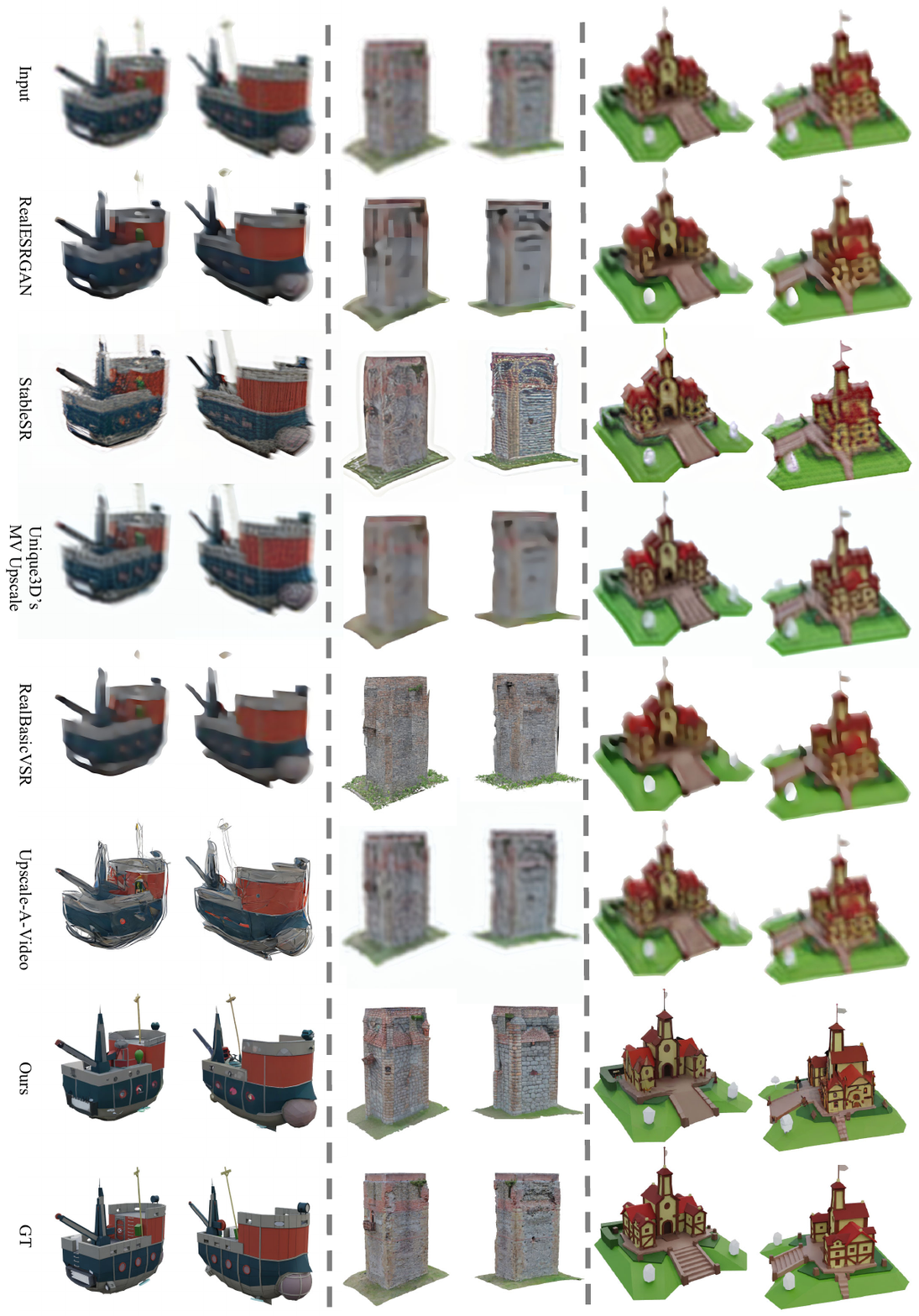}
    \vspace{-1mm}
    \caption{
        Qualitative comparisons on the Objaverse synthetic dataset. Our \NICKNAME{} demonstrates promising improvements, with increased detail and enhanced realism. (\textbf{Zoom in for best view.})
    }
    \vspace{-2mm}
\label{fig:qualitive_images_sys_supp}
\end{center}
\end{figure*}

\begin{figure*}[t]
\begin{center}
\includegraphics[width=.86\linewidth]{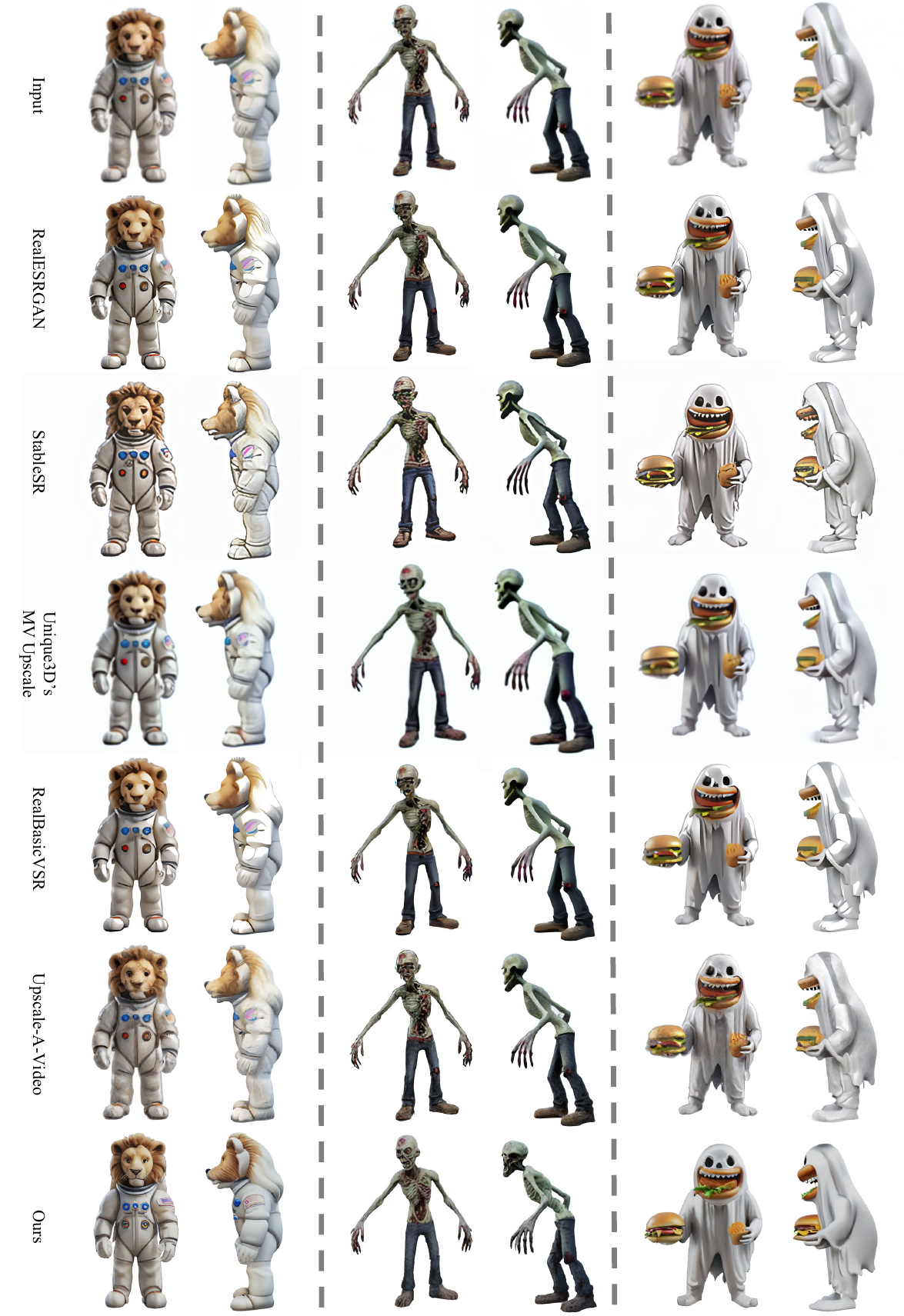}
    \vspace{-1mm}
    \caption{
        Qualitative comparisons on the in-the-wild
dataset. Our \NICKNAME{} yields significant improvements, providing enhanced detail and consistent output. (\textbf{Zoom in for best view.})
    }
    \vspace{-2mm}
\label{fig:qualitive_images_inthewild_supp}
\end{center}
\end{figure*}

In this section, we introduce another baseline from the multi-view image upscale module in Unique3D~\cite{wu2024unique3d}. This baseline fine-tunes ControlNet-Tile~\cite{zhang2023adding} to enhance RGB views. While the module can sharpen some textures, it struggles to recover inconsistent or corrupted areas in multi-view images. Our method outperforms Unique3D's MV Upscale both quantitatively and qualitatively. The quantitative comparisons between Unique3D's MV Upscale and our method is presented in Tab.~\ref{tab:unique3d}. Additionally, we provide more visual comparisons of our method with all other baselines, including RealESRGAN~\cite{wang2021realesrgan}, StableSR~\cite{wang2024stablesr}, Unique3D's MV Upscale~\cite{wu2024unique3d}, RealBasicVSR~\cite{chan2022realbasicvsr}, and Upscale-a-Video~\cite{zhou2024upscale}. Fig.~\ref{fig:qualitive_images_sys_supp} and Fig.~\ref{fig:qualitive_images_inthewild_supp} showcase the visual comparisons of multi-view enhancement on synthetic and in-the-wild datasets, respectively.

\begin{table}[h]
\caption{Quantitative comparisons of enhancing multi-view synthesis on the Objaverse synthetic dataset with Unique3D's MV Upscale module. Our method demonstrates clear advantages in restoration fidelity, as measured by PSNR, SSIM, and LPIPS. While applying color correction improves the output of Unique3D's MV Upscale module, it has minimal impact on our results when noise level is set to 0, highlighting our method's robustness against global color scheme shift issues.}
\centering
\vspace{-1mm}
\renewcommand{\arraystretch}{1.2}
\renewcommand{\tabcolsep}{2.05mm}
\resizebox{0.8\linewidth}{!} {
\begin{tabular}{c|cc|cc}
\toprule
Metrics & Unique3D's Upscale & Unique3D's Upscale (+ color fix) &  \textbf{\NICKNAME{}}  & \textbf{\NICKNAME{} (+ color fix)}\\ 
\midrule
 PSNR $\uparrow$   &  25.75  & 26.18 & 27.53 & 27.50 \\
 SSIM $\uparrow$  & 0.8989  & 0.9055  & 0.9265   &  0.9258 \\ 
 LPIPS $\downarrow$  & 0.1300  & 0.1257  & 0.0626   &  0.0631 \\ 
\bottomrule
\end{tabular}
}
\label{tab:unique3d}
\end{table}


\subsection{Video Demo}
\label{subsec:demo_video}
We also provide a demo video (\href{https://yihangluo.com/projects/3DEnhancer/#spotlight-video}{3DEnhancer-demo.mp4}) in our project page, showcasing visual results of  3D reconstruction enhancement.

\end{document}